\documentclass[preprint]{elsarticle}
\usepackage{graphicx} % Required for inserting images
\usepackage{siunitx}
\usepackage{amsmath, mathtools, amssymb, amsthm, geometry}
\newtheorem{remark}{Remark}
\usepackage{array, float}
\usepackage{verbatim}
\usepackage{mathrsfs}
\usepackage{svg}
\usepackage{algorithm}
\usepackage{algorithmic}
\usepackage{tikz}
\usepackage{pgfplots}
\usepackage{xcolor}
\usepgflibrary{fpu}
\usetikzlibrary{arrows.meta, positioning}
\usepackage{subcaption}
\usepackage[labelfont=bf,labelsep=period]{caption}
\usepackage[normalem]{ulem}
\newcommand{\yaw}{\psi}
\newcommand{\throt}{u}
\newcommand{\steer}{\delta}
\newcommand{\hl}[1]{{{\textcolor{orange}{[Hannah: #1]}}}}

\begin{document}

\begin{frontmatter}
\title{Data-Driven Modeling and Correction of Vehicle Dynamics}

\author[1]{Nguyen Ly}
\author[2]{Caroline Tatsuoka}
\author[3]{Jai Nagaraj}
\author[1]{Jacob Levy}
\author[1]{Fernando Palafox}
\author[1,4]{David Fridovich-Keil}
\author[1,4]{Hannah Lu}
\address[1]{Department of Aerospace Engineering and Engineering Mechanics, The University of Texas at Austin, Austin, TX 78712 USA}
\address[2]{Department of Mathematics, The Ohio State University, Columbus, OH 43210 USA}
\address[3]{Department of Computer Science, The University of Texas at Austin, Austin, TX 78712 USA}
\address[4]{Oden Institute for Computational Engineering and Sciences, The University of Texas at Austin, Austin, TX 78712 USA}

\begin{abstract}
We develop a data-driven framework for learning and correcting non-autonomous vehicle dynamics. Physics-based vehicle models are often simplified for tractability and therefore exhibit inherent model-form uncertainty, motivating the need for data-driven correction. Moreover, non-autonomous dynamics are governed by time-dependent control inputs, which pose challenges in learning predictive models directly from temporal snapshot data. To address these, we reformulate the vehicle dynamics via a local parameterization of the time-dependent inputs, yielding a modified system composed of a sequence of local parametric dynamical systems. We approximate these parametric systems using two complementary approaches. First, we employ the DRIPS (dimension reduction and interpolation in parameter space) methodology to construct efficient linear surrogate models, equipped with lifted observable spaces and manifold-based operator interpolation. This enables data-efficient learning of vehicle models whose dynamics admit accurate linear representations in the lifted spaces. Second, for more strongly nonlinear systems, we employ FML (Flow Map Learning), a deep neural network approach that approximates the parametric evolution map without requiring special treatment of nonlinearities. We further extend FML with a transfer-learning-based model correction procedure, enabling the correction of misspecified prior models using only a sparse set of high-fidelity or experimental measurements, without assuming a prescribed form for the correction term. Through a suite of numerical experiments on unicycle, simplified bicycle, and slip-based bicycle models, we demonstrate that DRIPS offers robust and highly data-efficient learning of non-autonomous vehicle dynamics, while FML provides expressive nonlinear modeling and effective correction of model-form errors under severe data scarcity.
\end{abstract}

\end{frontmatter}

\section{Introduction}
%%1. review about vehicle dynamics
Predicting the behavior of vehicular dynamical systems is both crucial and challenging due to the complex interactions between the vehicle, its control inputs, and its operating environment. A vehicle's trajectory depends not only on environmental factors such as terrain roughness, geometry, and deformability, but also on the vehicle's internal mechanisms, including steering configuration, actuation dynamics, suspension behavior, etc. \cite{iagnemma2002terrain,angelova2007learning,rogers2012continuous,lupu2024magic}. Constructing high-fidelity physics-based models that capture all of these effects is often infeasible or computationally prohibitive, motivating the use of simplified physics-based models that retain only the dominant modes of motion ~\cite{yu2021nonlinear, han2023model, westenbroek2023enabling}. However, these simplified models introduce approximation error, which is further compounded by variability in physical parameters and non-ideal components across different vehicles. Such discrepancies have motivated a growing body of work on data-driven learning of vehicle dynamics \cite{lee2023learning, gibson2023multi, dikici2025learning, levy2025meta}. Although these approaches improve predictive accuracy beyond what simplified physics models can provide, most of them struggle when high-fidelity data are scarce, and they rarely offer mechanisms for correcting existing physics-based models. These limitations underscore the need for data-driven methodologies that can effectively learn the dynamics and improve predictive performance, particularly in settings where time-dependent control inputs drive inherently non-autonomous dynamics.

%%2. review about data-driven modeling for non-autonomous dynamics
However, the input’s time-dependence, the defining feature of non-autonomous systems, complicates the data-driven learning process because it becomes difficult to distinguish intrinsic system behavior from externally driven variability, yielding an effectively infinite-dimensional parameter space~\cite{lu2024data,qin2021data}. For example, approaches based on dynamic mode decomposition (DMD) and Koopman operator theory attempt to capture non-autonomous dynamics by constructing linear models in lifted observable spaces, but accurately approximating time-varying Koopman operators remains challenging. Existing extensions, such as time-dependent Koopman modes~\cite{mezic2016koopman}, multi-resolution DMD~\cite{kutz2016multiresolution}, and delay-coordinate embeddings~\cite{das2019delay}, apply only to restricted classes of non-autonomous systems and typically require strong structural assumptions or large amounts of data~\cite{zhang2019online,macesic2018koopman}. Recent deep neural network-based approaches, such as Flow Map Learning (FML), learn the discrete-time evolution map directly~\cite{churchill2023flow}, but they still require a principled mechanism to handle time-dependent inputs. A key development in both DMD-based and DNN-based frameworks is the use of local parameterization of the external inputs~\cite{qin2021data,lu2024data}, which transforms a general non-autonomous system into a sequence of locally parametric autonomous systems whose dynamics can be learned from data. This strategy has enabled FML to approximate flow maps for general non-autonomous systems by learning locally parametric dynamics~\cite{qin2021deep}, and has further led to the deployment of Dimension Reduction and Interpolation in Parameter Space (DRIPS)~\cite{lu2023drips}, which builds on these ideas to achieve comparable predictive capability with orders-of-magnitude fewer training samples~\cite{lu2024data}. However, their performance on real physical systems has yet to be fully explored in realistic scenarios.

%% data-driven correction
Beyond learning non-autonomous dynamics from data, many practical applications, vehicle dynamics being a prime example, require correcting an existing physics-based model rather than replacing it entirely. Purely data-driven models trained on limited measurement data, common in real-world vehicle settings, can lack robustness, interpretability, and physical fidelity, often leading to extrapolation errors or unrealistic predictions \cite{wang2024pay, levylearning}. Model correction techniques often assume an additive and/or a multiplicative correction term; these terms have been parameterized by Gaussian Process (GP) \cite{kennedy2001bayesian, higdon2004combining}, stochastic expansions \cite{eldred2017multifidelity,sargsyan2015statistical}, and neural networks \cite{chen2021generalized,zou2024correcting}. Recent developments in data-driven model correction address this challenge by leveraging abundant low-fidelity data from a prior model (i.e., the simplified physics-based model with prior estimate of model parameters) together with a scarce set of high-fidelity or experimental observations. A representative approach constructs an FML approximation of the prior model. Then it fine-tunes the model using high-fidelity data, thereby transferring structural information encoded in the low-fidelity model while correcting systematic discrepancies~\cite{tatsuoka2025deep}. This transfer-learning-based strategy has been shown to improve predictive accuracy even when high-fidelity data are extremely limited, without requiring an explicit form of the correction term. However, the effectiveness of this correction technique has not been explored for non-autonomous vehicle systems, especially in settings where corrections must be derived from real measurement data, which motivates the developments presented in this work.

%%3. our contribution
In this work, we apply the recent methodology developments on data-driven learning and correcting to non-autonomous vehicle dynamics, addressing key challenges in both data efficiency and model fidelity. By parameterizing the time-dependent control inputs locally, we transform the original non-autonomous dynamics into a sequence of locally parametric autonomous systems, enabling effective data-driven learning. Building on this formulation, we employ DRIPS to construct data-efficient surrogate models via DMD and manifold-based operator interpolation, allowing accurate prediction of vehicle trajectories under unseen control inputs. For more strongly nonlinear behaviors, we leverage FML and extend it with a transfer-learning-based correction that incorporates scarce high-fidelity or real measurement data to refine prior models without prescribing an explicit correction form. We demonstrate the capability of this framework on unicycle, simplified bicycle, and slip-based bicycle dynamics, showing that DRIPS provides robust and data-efficient learning of vehicle dynamics, while the FML approach captures complex nonlinear effects and significantly improves prediction accuracy when prior models are corrected from transfer learning. Altogether, this work provides a cohesive strategy for both learning and correcting vehicle dynamical models in realistic, time-varying settings, bridging idealized physics-based models with real-world measurement data through the advantages of modern data-driven tools.

%%4. structure of the paper
The remainder of this paper is organized as follows. Section~\ref{sec:dynamicsOverview} describes the physics-based vehicle models used in this study, based on approximations of a NVIDIA JetRacer platform equipped with two-wheel drive and differential steering. Section~\ref{sec:dataDriven} presents the theoretical foundations of the proposed framework, including the DRIPS methodology and the transfer-learning-based extension of flow-map learning. Section \ref{sec:num-ex} details the training and validation procedures and presents a comprehensive analysis of the resulting predictions across multiple vehicle models, demonstrating strong accuracy and generalization under diverse control inputs. Section~\ref{sec:conclusion} concludes the paper by summarizing the main findings and outlining potential directions for extending these data-driven approaches to more complex vehicle dynamics.

\section{Vehicle Dynamics}
\label{sec:dynamicsOverview}
The dynamical behavior of a vehicle can be expressed in the form of a non-autonomous ordinary differential equation (ODE),
\begin{subequations}\label{eq:general}
\begin{equation}
\frac{d\mathbf{s}(t)}{dt} = \mathbf{f}_\text{true}(\mathbf{s}(t), \mathbf{c}(t)), \quad t \in (0, T],
\end{equation}
where $\mathbf{s} = [s_1,\cdots, s_{N_s}]^\top\in\Omega_{\mathbf s}\subset \mathbb{R}^{N_s}$ is the state of the system and $\mathbf{c} = [c_1,\cdots,c_{N_c}]^\top\in \Omega_{\mathbf c}\subset\mathbb{R}^{N_c}$ represents a vector of arbitrary control inputs defined by the user. The system is typically initialized with
\begin{equation}
    \mathbf{s}(0)=\mathbf{s}^0,
\end{equation}
\end{subequations}
and its evolution is governed by the functional form of $\mathbf{f}_\text{true}$, which can be linear or nonlinear. In practice, however, the exact dynamics are often unknown or too complex to model directly. Consequently, \textit{simplified approximations} are introduced to capture the dominant behaviors while maintaining computational tractability. We denote such models as $\mathbf f_\text{prior}$, representing our prior knowledge of the underlying system, i.e.,
\begin{equation}\label{eq:prior}
\frac{d\mathbf{s}(t)}{dt} = \mathbf{f}_\text{prior}(\mathbf{s}(t), \mathbf{c}(t)).
\end{equation}

This paper will consider three such simplified representations: the unicycle model (Section~\ref{sec:unicycle}), the ``simplified" or ``slip-free" planar bicycle model (Section~\ref{sec:noSlip}), and the ``slip-based" planar bicycle model (Section~\ref{sec:yesSlip}). These models, which are widely used in vehicle dynamics and control, form a hierarchy of increasing complexity and fidelity. Each model establishes certain assumptions about the interaction between the vehicle and the environment that reduce the number of state variables being considered. While these physics-based models provide interpretability and physical consistency, their predictive accuracy depends on parameters, such as tire stiffness and mass distributions, that are often uncertain and/or time-varying. This model-form uncertainty motivates the integration of data-driven modeling and correction, which can learn and compensate for unmodeled dynamics, thereby improving prediction fidelity and precise control.

\subsection{Unicycle Dynamics}
\label{sec:unicycle}
The unicycle model is a set of equations governing the behavior of the namesake system it describes: a unicycle. It describes the kinematic behavior of an idealized vehicle that moves on a plane without lateral slip. Using only the forward velocity in the body frame, $v(t)$, and the yaw angle rate, $\omega(t)$, as control inputs, the model governs the evolution of three state variables: the horizontal position $x(t)$ and the vertical position $y(t)$ in the world frame, and the yaw angle $\yaw(t)$ that defines the vehicle's heading. These can be compactly expressed as \(\mathbf{s}(t) = [x(t),\, y(t),\, \yaw(t)]^\top\) and \(\mathbf{c}(t) = [v(t),\, \omega(t)]^\top
\). The corresponding equations of motion are given by 
\begin{subequations}\label{eq:unicyle}
\begin{align}
    &\frac{d}{dt}x(t) = v(t)\cos(\yaw(t)), \\
    &\frac{d}{dt}y(t) = v(t)\sin(\yaw(t)), \\
    &\frac{d}{dt}\yaw(t) = \omega(t).
\end{align}
\end{subequations}
Here, the body frame is a coordinate system whose origin is attached to the vehicle's center of mass; it accelerates and rotates relative to the inertial frame of reference. Planar (two-dimensional) motion is viewed from a top-down perspective, where the observer looks along the negative $z$-axis only and the body frame $x$-$y$ plane rotates. The yaw angle $\yaw$ characterizes this rotation, measured counterclockwise from the inertial $x$-$y$-axes. Figure \ref{fig:unicycleEx} shows the unicycle model and its response to a selected control set:
\begin{figure}[H]
    \centering
    \includegraphics[width=\linewidth]{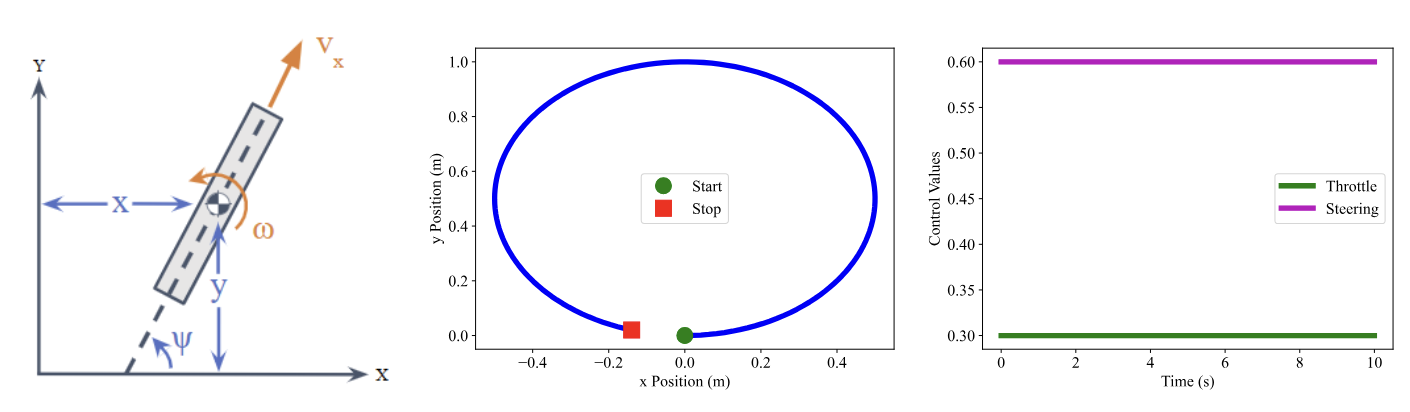}
    
    \caption{Left: unicycle model; Middle: trajectory under constant forward velocity and yaw rate; Right: control inputs over time.}
    \label{fig:unicycleEx}
\end{figure}

The unicycle model assumes that the vehicle's instantaneous velocity vector is perfectly aligned with its heading direction, implying zero lateral slip. While this assumption limits its validity during higher speeds or sharp maneuvers, the model remains widely used in motion planning and control due to its analytical simplicity and geometric interpretability. It also provides a useful baseline for developing and evaluating our data-driven modeling framework.

\subsection{Planar Bicycle Dynamics}
The planar bicycle model provides a higher-fidelity representation of vehicle motion by accounting for steering geometry and, in its extended form, tire slip dynamics. Although originally derived for bicycles, the same formulation can approximate any two- and four-wheeled vehicle whose left and right wheels are assumed to share the same lateral velocity and slip angle. Similar to the unicycle model (Section \ref{sec:unicycle}), the system is controlled through two inputs: a longitudinal command $\throt(t)$ and a steering command $\steer(t)$, i.e., $\mathbf c(t) = [\throt(t),\steer(t)]^\top$. In this study, both inputs are represented as normalized voltage signals in the range $[-1, 1]$, which are then scaled by \textit{scaling coefficients} $b_\throt$ and $b_\steer$ to yield physical units $\text{m/s}^2$ and $\text{rad}$, respectively.

\subsubsection{Slip-Free Model}\label{sec:noSlip}
\begin{figure}[H]
    \centering
    \includegraphics[width=\linewidth]{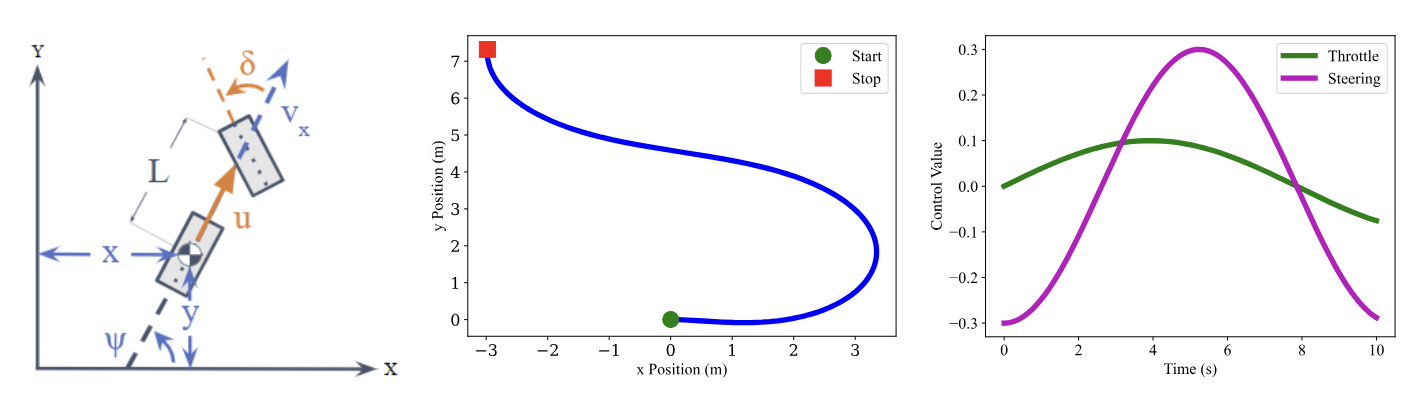}

    \caption{Left: simplified planar bicycle model; Middle: trajectory under sinusoidal throttle and steering inputs; Right: control inputs over time.}
    \label{fig:bicycleNoSlipEx}
\end{figure}

The simplified or \textit{slip-free} bicycle model describes planar motion using four state variables: horizontal position $x(t)$ and vertical position $y(t)$ in the world frame, the forward velocity $v_x(t)$ in the body frame, and the yaw angle $\yaw(t)$, i.e., $\mathbf s(t) = [x(t),y(t),v_x(t),\yaw(t)]^\top$. Assuming no tire slip, the equations of motion are given by
\begin{subequations}\label{eq:simp-bicyle}
\begin{align}
    &\frac{d}{dt}x(t) = v_x(t)\cos(\yaw(t)), \\
    &\frac{d}{dt}y(t) = v_x(t)\sin(\yaw(t)), \\
    &\frac{d}{dt}v_x(t) = b_\throt\throt (t),\\
    &\frac{d}{dt}\yaw(t) = \frac{v_x(t)}{L} \tan(b_\steer \steer(t)),
\end{align}
\end{subequations}
where $L$ denotes the wheelbase, or the distance between the front and rear axles. Figure \ref{fig:bicycleNoSlipEx} shows the simplified bicycle model and its response to a sinusoidal control signal.

This model assumes that the tires maintain pure rolling contact with the ground, meaning that the direction of motion at the contact patch is always tangent to the wheel plane. As a result, lateral tire forces are neglected, and the instantaneous center of rotation lies along the rear axle line. These assumptions make the model suitable for low-speed maneuvers, such as parking or trajectory tracking, where tire slip angles are small and inertial effects can be ignored.

The simplified bicycle model strikes a balance between the geometric simplicity of the unicycle model and the physical realism of the full dynamic model. It introduces steering geometry through the wheelbase and steering angle, enabling a more accurate representation of turning behavior while remaining computationally efficient. However, because it neglects lateral dynamics and load transfer effects, it becomes less accurate at higher speeds or during aggressive steering. The slip-based formulation in Section~\ref{sec:yesSlip} relaxes these assumptions to capture more realistic vehicle behavior.

\subsubsection{Slip-Based Model}\label{sec:yesSlip}
\begin{figure}[H]
    \centering
    \includegraphics[width=\linewidth]{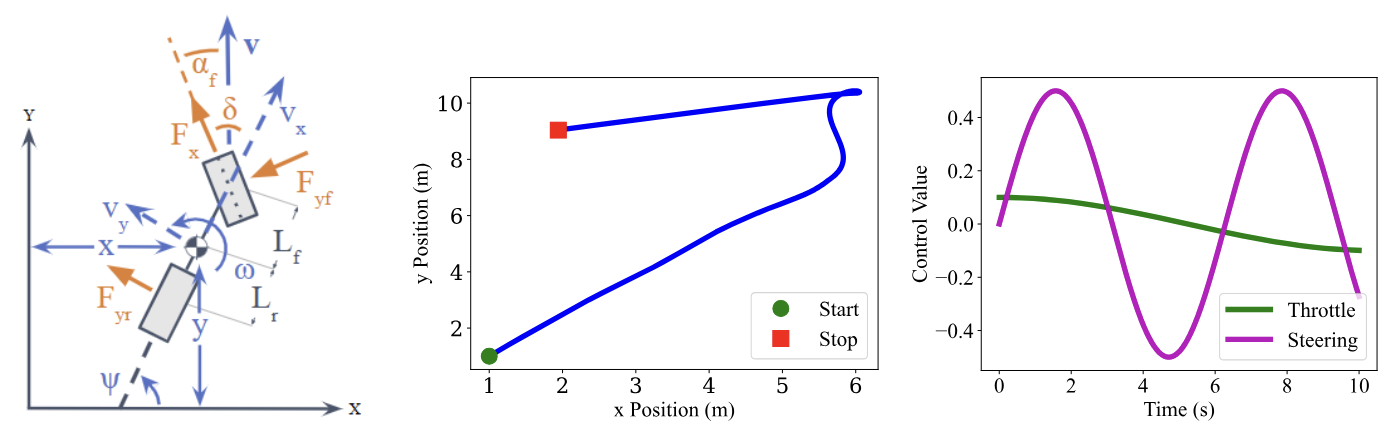}
    
    \caption{Left: slip-based planar bicycle model; Middle: trajectory under sinusoidal throttle and steering inputs; Right: control inputs over time.}
    \label{fig:bicycleSlipEx}
\end{figure}

The \textit{slip-based} bicycle model augments the simplified formulation by accounting for lateral motion and rotational dynamics of the vehicle body. Two additional state variables are introduced: the lateral velocity $v_y(t)$ in the body frame and the yaw angle rate $\omega(t)$, i.e., $\mathbf s(t) = [x(t),y(t),v_x(t),\yaw(t),v_y(t),\omega(t)]^\top$. These quantities describe side-slip and rotational motion, respectively, and enable the model to represent dynamic behaviors such as drifting, understeer, and oversteer. The control inputs remain as $\mathbf c(t) = [\throt(t),\steer(t)]^\top$, but their effects are transmitted through tire forces rather than directly influencing kinematic variables. The governing equations are expressed as
\begin{subequations}\label{eq:full-bicyle}
\begin{align}
    &\frac{d}{dt}x(t) = v_x(t)\cos(\yaw(t)) - v_y(t)\sin(\yaw(t)), \\
    &\frac{d}{dt}y(t) = v_x(t)\sin(\yaw(t)) + v_y(t)\cos(\yaw(t)), \\
    &\frac{d}{dt}\yaw(t) = \omega(t), \\
    &\frac{d}{dt}v_x(t) = \frac{1}{m}\big(F_x\cos(b_\steer \steer(t))-F_{yf}\sin(b_\steer \steer(t))\big) - \omega(t) v_y(t),\\
    &\frac{d}{dt}v_y(t) = \frac{1}{m}\big(F_{yf}\cos(b_\steer \steer(t)) + F_x\sin(b_\steer \steer(t)) + F_{yr}\big) - \omega(t) v_x (t),\\
    &\frac{d}{dt}\omega(t) = \frac{1}{I_z}\Big(L_f\big(F_{yf}\cos(b_\steer \steer(t)) + F_x \sin(b_\steer \steer(t))\big) - L_rF_{yr}\Big),
\end{align}
\end{subequations}
where $m$ is the vehicle mass and $I_z$ is the yaw-axis moment of inertia. The wheelbase parameter $L$ from~\eqref{eq:simp-bicyle} is now divided into $L_f$ and $L_r$ to represent the distances from the center of mass to the front and rear wheels, respectively. We assume that the slip occurs only in the lateral direction of the wheels, so that the forward tire force $F_x$ is given by:
\begin{equation}\label{eq:forwardForce}
    F_x(  \throt(t)) = mb_\throt \throt(t).
\end{equation}
The lateral forces are modeled using a linearized relationship between lateral force and slip angle. For small slip angles $\alpha_f\approx\alpha_r\approx\frac{-v_y}{v_x}$, the front and rear lateral forces are approximated by 
\begin{subequations}\label{eq:lateralForce}
\begin{align}
&F_{yf}( \steer(t), v_x(t),v_y(t),\omega(t)) = -C_f \left( b_\steer\steer(t) - \frac{v_y(t) + L_f\omega(t)}{v_x(t)} \right), \\
&F_{yr}(v_x(t),v_y(t),\omega(t)) = -C_r \left(\frac{v_y(t) + L_r\omega(t)}{v_x(t)} \right),
\end{align}
\end{subequations}
where $C_f$ and $C_r$ are the cornering stiffness of the front and rear tires, respectively. Figure \ref{fig:bicycleSlipEx} shows the slip-based bicycle model and its response to a control set.

By incorporating these additional effects, the slip-based model captures the coupling between longitudinal, lateral, and yaw motions that dominate vehicle dynamics at higher speeds. It provides a more physically realistic description of tire-road interaction, enabling the study of stability, handling, and dynamic control. However, the model also introduces parameters, such as tire stiffness, friction coefficients, and load transfer effects, which are difficult to measure and often vary with operating conditions. These challenges motivate the use of data-driven correction techniques, which will be discussed in Section~\ref{sec:correction}.

\section{Data-Driven Models of Non-autonomous Systems}\label{sec:dataDriven}
While these physics-based formulations~\eqref{eq:unicyle}-\eqref{eq:full-bicyle} offer interpretable, both analytically and computationally tractable structure,
% and analytical tractability, 
their predictive capability often deteriorates when the system operates under time-varying or uncertain conditions not captured by fixed parameters or simplified constitutive laws. In practice, vehicle dynamics are inherently non-autonomous, driven by changing control inputs, environmental factors, and interactions that evolve. To effectively model such systems, data-driven approaches have emerged as a powerful alternative, learning the governing dynamics directly from observational or simulated data.

Given a set of $N_\text{obs}$ temporal snapshots of the state variables $\mathbf{s}(t)$ collected at times $t_{obs}=\{t_1, \dots,t_{N_\text{obs}}\}$ from the true unknown system~\eqref{eq:general}, our goal is to construct a numerical surrogate model $\hat{\mathbf{s}}(t)$ from these snapshots such that it approximates the true state trajectory $\mathbf{s}(t)$ accurately over the domain in (\ref{eq:general}) for any initial conditions $\mathbf{s}^0$ and input $\mathbf{c}(t)$. In mathematical terms: 
\begin{equation}\label{defSurrogate}
\hat{\mathbf s}(t_k;\mathbf s_0,\mathbf{c}(t_k))\approx \mathbf s(t_k;\mathbf s_0,\mathbf{c}(t_k)),\qquad k = 1,\dots,N_T,\quad 0 = t_0 < \dots < t_{N_T} = T.
\end{equation}
To simplify the presentation and without loss of generality, we consider a uniform time discretization $t_k = k\Delta t\in [0,T]$ with $k = 0,\dots, N_T$.
%% I wonder if using $\delta t_k$ is viable in making a concise report but also capturing the generality of the discrete representation below.

Our data-driven framework for learning \eqref{eq:general} consists of two major steps: first, we decompose the dynamical system into a sequence of local systems by parameterizing the external input $\mathbf{c}(t)$ locally in time; second, we learn the local parametric systems via data-driven methods such as DRIPS and FML. 

\subsection{Local parameterization and modified system}\label{localParam}
A discrete-time representation of~\eqref{eq:general} is 
\begin{equation}\label{eq:HFM-dis}
\mathbf s(t_{k+1}) = \mathbf{f}^{\Delta t}_\text{true}(\mathbf s(t_{k}),\mathbf{c}(t_k)):=\mathbf s(t_{k})+\int_{0}^{\Delta t} \mathbf{f}_\text{true}(\mathbf s(t_k+\tau),\mathbf{c}(t_k+\tau)) \text d\tau,
\end{equation}
where $t_k\in [0,T]$. In each time interval $[t_k, t_{k+1}]$, $k = \{0,\dots, N_T-1\}$, we use a \textit{local} parameterization of $\mathbf{c}(t)$ in the form similar to~\cite{qin2021data}: 
\begin{equation}\label{eq:local_par}
\tilde{\mathbf{c}}_k(\tau; \mathbf{p}_k):= \sum_{j=1}^{N_\text{par}}p_k^j b_j(\tau)\approx \mathbf{c}(t_k+\tau),\qquad \tau \in [0,\Delta t],
\end{equation}
where $b_j(\tau)$ with $j = \{1,\dots, N_\text{par}\}$ is a set of $N_\text{par}$ prescribed analytical basis functions, and 
\begin{equation}\label{eq:par}
\mathbf p_k = [p_k^1,\cdots, p_k^{N_\text{par}}]^\top\in\Omega_{\mathbf p}\subset \mathbb R^{N_\text{par}}
\end{equation}
are the basis coefficients parameterizing $\mathbf{c}(t)$ in $[t_k,t_{k+1}]$. Examples of local parameterization techniques include interpolating polynomials and Taylor polynomials (cf. Section 3.1 in~\cite{qin2021data}). Then, a \textit{global} parameterization of $\mathbf{c}(t)$ is constructed as 
\begin{subequations}\label{eq:global_par}
\begin{equation}
\tilde{\mathbf{c}}(t;\mathbf p)= \sum_{k = 0}^{N_T-1}\tilde{\mathbf{c}}_k(t-t_k;\mathbf p_k)\mathbb I_{[t_k,t_{k+1}]}(t), \qquad \mathbf p := \{\mathbf p_k\}_{k=0}^{N_T-1}\in  \mathbb R^{N_T\times N_\text{par}},
\end{equation}
where $\mathbf p$ is a global parameter set for $\tilde{\mathbf{c}}(t)$, and $\mathbb I_{[a,b]}$ is the indicator function 
\begin{equation}
\mathbb I_{[a,b]} (t) = \left\{
\begin{aligned}
&1&&\text{if }t\in [a,b],\\
&0&&\text{otherwise}.
\end{aligned}
\right.
\end{equation}
\end{subequations}

Now we consider a modified representation of the true (unknown) system~\eqref{eq:general}, 
\begin{equation}\label{eq:modified}
\left\{
\begin{aligned}
&
\frac{d\tilde {\mathbf{s}}(t)}{dt} = \mathbf{f}_\text{true}(\tilde{\mathbf{s}}(t), \tilde{\mathbf{c}}(t;\mathbf p)), \quad t \in (0, T],\\
&\tilde{\mathbf{s}}(0) = \mathbf s_0,
\end{aligned}
\right.
\end{equation}
where $\tilde{\mathbf c}(t;\mathbf p)$ is the globally parameterized input defined in~\eqref{eq:global_par}. When the control input $\mathbf c (t)$ is already known or given in a parametric form, i.e., when $\tilde {\mathbf c} (t) = \mathbf c (t)$, the modified system~\eqref{eq:modified} is equivalent to the original system~\eqref{eq:general}. When the parameterized process $\tilde {\mathbf c} (t)$ needs to be numerically constructed, the modified system~\eqref{eq:modified} becomes an approximation to the true system~\eqref{eq:general}. The approximation accuracy depends on the accuracy of $\tilde {\mathbf c}(t)\approx \mathbf c(t)$.

According to Lemma 3.1 in~\cite{qin2021data}, there exists a function $\tilde {\mathbf f}^{\Delta t} : \Omega_{\mathbf s}\times  \Omega_{\mathbf p}\to  \Omega_{\mathbf s}$, which depends on $ \mathbf f_\text{true}$, such that, for any time interval $[t_k, t_{k+1}]$, the solution of \eqref{eq:modified} satisfies
\begin{equation}\label{eq:modified-HFM-dis}
\tilde{\mathbf s}_{k+1} = \tilde { \mathbf f}^{\Delta t} (\tilde{\mathbf s}_{k},\mathbf p_k),\qquad k = 0,\dots, N_T-1,
\end{equation}
where $\mathbf p_k$ is the local parameter set in~\eqref{eq:par} for the locally parameterized input $\tilde {\mathbf c}_k(t)$ in~\eqref{eq:local_par}. Note that \eqref{eq:modified-HFM-dis} provides the evolution of the solution, and the time variable is eliminated.

\subsection{Data-Driven Discovery of the modified system}
The function $\tilde{\mathbf f}^{\Delta t}$ in~\eqref{eq:modified-HFM-dis} governs the evolution of the solution to the modified system~\eqref{eq:modified} and is the target function to learn. The challenge posed by non-autonomous systems is now shifted to the task of learning the parametric system~\eqref{eq:modified-HFM-dis} in any time interval $[t_k,t_{k+1}]$. This task can be completed by DRIPS or FML.

\subsubsection{Training and testing datasets}
\label{sec:dataset}

Consider a set of $N_\text{sam}$ samples of the model input $\{\mathbf c^{(1)}(t), \dots, \mathbf c^{(N_\text{sam})}(t)\}$. Each sample is evaluated at discrete times $0 = t_0 < t_1 < \dots t_k < \dots < t_{N_{T}} = T$ with $\Delta t = t_{k+1}-t_k$, $k = 0,\dots, N_T-1$.
For the $i^\text{th}$ sample, we arrange these inputs and system responses as
\[
\text{input}: \{ \mathbf s^{(i)}(t_k), \mathbf c^{(i)}(t_k) \} ,\qquad \text{output}: \{ \mathbf s^{(i)}(t_{k+1}) \}
\]
pairs, and treat them as representative of the true discrete-time dynamical system~\eqref{eq:HFM-dis} in the time interval $[t_k,t_{k+1}]$, i.e.,
\begin{equation}\label{eq:flow_map_true}
\mathbf s^{(i)}(t_{k+1}) = \mathbf f^{\Delta t}_\text{true}(\mathbf s^{(i)}(t_k), \mathbf c^{(i)}(t_k)),\qquad 
k = 0,\dots N_T-1, \quad i = 1,\dots, N_\text{sam}.
\end{equation}

The local parameterization of $\mathbf c^{(i)}(t_k)$ gives $\tilde {\mathbf c}_k^{(i)}(\tau;\mathbf p_k^{(i)})$, where $\tau\in [0, \Delta t]$ and $\mathbf p_k^{(i)}$ is the parameter set for the local parameterization of the input in the form of~\eqref{eq:local_par}. Along the $i^\text{th}$ sample trajectory and during the $k^\text{th}$ time interval, a local dataset  is
\begin{equation}
\mathcal S_\text{train}^{(k,i)} = \Big\{\{\mathbf s^{(i)}_k,\mathbf p_k^{(i)}\} \; , \; \{\mathbf s^{(i)}_{k+1}\}\Big\}.
\end{equation}
These input/output pairs satisfy  the modified system~\eqref{eq:modified-HFM-dis}, 
\begin{equation}\label{eq:train-HFM}
\mathbf s^{(i)}_{k+1} \approx \tilde { \mathbf f}^{\Delta t} (\mathbf s^{(i)}_{k},\mathbf p^{(i)}_k),
\qquad k = 0,\dots N_{T}-1, \quad i = 1,\dots, N_\text{sam}.
\end{equation}
We assemble these data into the training dataset 
\begin{equation}\label{eq:training_data}
\mathcal S_\text{train} = \bigcup\limits_{k,i} \mathcal S_\text{train}^{(k,i)}, \qquad k = 0,\dots N_{T}-1, \qquad i = 1,\dots, N_\text{sam},
\end{equation}
for the full simulation-time horizon $t \in [0,T]$. As we observe from~\eqref{eq:train-HFM}, the time variable does not play an explicit role in the learning process. So, to better learn the map $\tilde{\mathbf f}^{\Delta t}: \Omega_{\mathbf s}\times \Omega_{\mathbf p}\to \Omega_{\mathbf s}$ in practice, a preferred dataset can be generated from sampling over the space $\Omega_{\mathbf s}\times \Omega_{\mathbf p}$ as follows
\begin{subequations}\label{eq:train_data}
\begin{equation}
\mathcal S_\text{train} = \bigcup\limits_{m,j} \mathcal S_\text{train}^{(m,j)},  \qquad m = 0,\dots N_\text{sam}^{\mathbf s}, \qquad j = 1,\dots, N_\text{sam}^{\mathbf p},
\end{equation}
where 
\begin{equation}\label{eq:train_data_sub}
\mathcal S_\text{train}^{(m,j)} = \{\{\mathbf s^{(m,j)}_\text{in},\mathbf p^{(j)}\} \; , \; \{\mathbf s^{(m,j)}_\text{out}\}\},
\end{equation}
\end{subequations}
and the input/output pairs satisfy approximately
\begin{equation}\label{eq:train-HFM2}
\mathbf s^{(m,j)}_\text{out} \approx \tilde { \mathbf f}^{\Delta t} (\mathbf s^{(m,j)}_\text{in},\mathbf p^{(j)}),\qquad m = 0,\dots N_\text{sam}^{\mathbf s},
 \qquad j = 1,\dots, N_\text{sam}^{\mathbf p}.
\end{equation}
The sampling strategy in $\Omega_{\mathbf s}$ and $\Omega_{\mathbf p}$ will be specified in Section~\ref{sec:num-ex} for each problem.

Our goal is to build, from the dataset $\mathcal S_\text{train}$, a surrogate model of the (unknown) system~\eqref{eq:modified-HFM-dis}. This surrogate should yield a low-cost prediction of the true system-state dynamics~\eqref{eq:general}, 
\begin{equation}\label{eq:data_test}
\mathcal S_\text{test} = \{\mathbf s(t_0), \dots, \mathbf s(t_{N_{T^*}}) \},
\end{equation}
for an arbitrary input $\mathbf c^*(t)$ not seen during the training: $\mathbf c^*(t)\notin \{\mathbf c^{(i)}(t)\}_{i=1}^{N_\text{sam}}$. Another goal is to use this surrogate to make predictions over the discrete time instances
\begin{equation}\label{eq:test_time}
    0=t_0<t_1<\cdots<t_k<\cdots<t_{N_{T^*}} = T^*,
\end{equation}
where $T^*$ can be very large as the evolution~\eqref{eq:modified-HFM-dis} does not depend on $t$ explicitly. 

\subsubsection{DRIPS}
\label{drips}
Dimension Reduction and Interpolation in Parameter Space (DRIPS) is a two–stage data-driven framework for discovering the modified parametric dynamics $\tilde{\mathbf f}^{\Delta t}(\mathbf s,\mathbf p)$.
\begin{itemize}
\item \textbf{Training (offline) stage:} DRIPS first learns \emph{local} linear flow maps $\mathbf L(\mathbf p^{(j)})$ that best fit the local training sets $\bigcup_m\mathcal S_\text{train}^{(m,j)}$ at local parameter points $\mathbf p^{(j)}$ for $j=1,\cdots,N_\text{sam}^{\mathbf p}$. The linear operator $\mathbf L(\mathbf p^{(j)})$ is typically represented by a local parametric reduced-order model (PROM) $\mathbf L_r(\mathbf p^{(j)})$ of rank $r$ with a corresponding local reduced order basis (ROB) $\mathbf V(\mathbf p^{(j)})$. 
\item \textbf{Predicting (online) stage:} Given a new control input $\mathbf c^*(t)$, we use local parameterization~\eqref{eq:local_par} to approximate $\mathbf c^*(t_k)$ with $\tilde{\mathbf c}_k^*(\tau;\mathbf p_k^*)$ for $k = 0,\cdots, N_{T^*}$. DRIPS then \emph{interpolates} the learned operators $\{\mathbf L(\mathbf p^{(j)})\}_{j=1}^{N_\text{sam}^{\mathbf p}}$ across the parameter domain $\Omega_{\mathbf p}$ to construct a fast parametric model, yielding a sequence of linear flow maps $\{\mathbf L(\mathbf p_k^*)\}_{k=0}^{N_{T^*}}$ for new parameter queries $\{\mathbf p_k^*\}_{k=0}^{N_{T^*}}$. The construction of each $\mathbf L(\mathbf p_k^*)$ is obtained by computing the corresponding $\mathbf L_r(\mathbf p_k^*)$ and $\mathbf V(\mathbf p_k^*)$ from interpolating $\{\mathbf L_r(\mathbf p^{(j)})\}_{j=1}^{N_\text{sam}^{\mathbf p}}$ and $\{\mathbf V(\mathbf p^{(j)})\}_{j=1}^{N_\text{sam}^{\mathbf p}}$ respectively.
\end{itemize}

The overall framework is summarized in Algorithm~\ref{alg:drips}, where $\mathcal{I}_V$ and $\mathcal{I}_L$ denote, respectively, the interpolation of ROBs on the Grassmann manifold~\cite{absil2004riemannian,boothby2003introduction,helgason1979differential,rahman2005multiscale,edelman1998geometry} and the interpolation of PROMs on the matrix manifold~\cite{amsallem2008interpolation,amsallem2011online}. Both interpolations are nontrivial and draw on the extensive literature on Proper Orthogonal Decomposition (POD)-based reduced-order modeling~\cite{amsallem2008interpolation,amsallem2011online}. At each training parameter point $\mathbf p^{(j)}$, the Dynamic Mode Decomposition (DMD) or Koopman-augmented DMD is employed to identify a low-dimensional linear representation of the local dynamics $\tilde {\mathbf f}^{\Delta t}(\mathbf s, \mathbf p^{(j)})$.

Given snapshot pairs $\bigcup\limits_{m}\mathcal S_\text{train}^{(m,j)}$, the method seeks a linear operator $\mathbf L(\mathbf p^{(j)})$ that best approximates the evolution map 
\begin{equation}
\mathbf s^{(m,j)}_\text{out}\approx \mathbf L(\mathbf p^{(j)}) \mathbf s^{(m,j)}_\text{in}\quad \text{ for all }\quad m = 0,\cdots, N_\text{sam}^{\mathbf s}.
\end{equation}
In practice, the rank-$r$ reduced operator $\mathbf L_r(\mathbf p^{(j)}) = \mathbf V(\mathbf p^{(j)})^\top\mathbf L(\mathbf p^{(j)})\mathbf V(\mathbf p^{(j)})$ and the corresponding ROB $\mathbf V(\mathbf p^{(j)})$ are used for efficient computation and dimensionality reduction. The Koopman-augmented formulation~\cite{kutz2016dynamic,tu2013dynamic} enriches the state observables by an invertible nonlinear feature lifting maps $\mathbf g$, allowing the resulting reduced operator to capture nonlinearity in the original dynamics while preserving a linear evolution in the lifted space: 
\begin{equation}
\mathbf g(\mathbf s^{(m,j)}_\text{out})\approx \mathbf L(\mathbf p^{(j)})\mathbf g(\mathbf s^{(m,j)}_\text{in})\quad \text{ for all }\quad m = 0,\cdots, N_\text{sam}^{\mathbf s}.
\end{equation}
Then, in prediction mode, the state is advanced recursively by the interpolated operator according to
\begin{equation}\label{eq:DRIPS-pred}
\left\{
\begin{aligned}
    &\hat{\mathbf s}(0) = \mathbf s(0),\\
    &\hat{\mathbf s}(t_{k+1})=\mathbf L(\mathbf p^*_k) \hat{\mathbf s}(t_k)
    \end{aligned}
    \right.
    \quad \text{or}    \quad 
    \left\{
    \begin{aligned}
    &\hat{\mathbf s}(0) = \mathbf s(0),\\
    &\hat{\mathbf s}(t_{k+1})=\mathbf g^{-1}\left(\mathbf L(\mathbf p^*_k) \mathbf g(\hat{\mathbf s}(t_k))\right)
    \end{aligned}
    \right.\quad \text{ for } k = 0,\cdots, N_{T*}-1.
\end{equation}
The choice of the lifting map $\mathbf g$ is highly problem-dependent and should be guided by physical insight into the governing processes, e.g., \cite{lu2020lagrangian,lu2020prediction,lu2021dynamic,lu2021extended}. A choice of the observables may not be unique and optimal. The task of optimally constructing observables is beyond the scope of this paper; instead, we employ several standard techniques in DMD studies to construct reasonable observables, as illustrated in the numerical example sections. Further details of the DRIPS algorithm and its theoretical background can be found in~\cite{lu2023drips, lu2024data}.

\begin{algorithm}[h!]
\caption{DRIPS framework (adapted from~\cite{lu2023drips,lu2024data})}
\label{alg:drips}
\vspace{1mm}
\textit{Offline Stage:} 
\vspace{1mm}

\textbf{For} $j = 1,\dots,N_\text{sam}^{\mathbf p}$,
$$
\begin{aligned}
&\text{Get the training data~\eqref{eq:train_data}, } \\
&\text{Input:} \ \bigcup\limits_{m}\mathcal S_\text{train}^{(m,j)} \xrightarrow[\text{training}]{\text{DMD or Koopman augmented DMD}} \text{Output:} \ \mathbf V(\mathbf p^{(j)}) \ \text{and}  \ \mathbf L_r(\mathbf p^{(j)});
\end{aligned}$$

\textbf{End}\\

\textit{Online Stage:}
\vspace{1mm}

\textbf{Input:} $\hat{\mathbf s}(0) = \mathbf s(0)$ and $\{\mathbf p_k^*\}_{k=0}^{N_{T^*}-1}$ obtained from local parameterization~\eqref{eq:local_par} of the test control input $\mathbf c^*(t)$ over prediction time horizon $[0, T^*]$.
\vspace{1mm}

\textbf{For} $k = 0,\dots,N_{T^*}-1$,

\begin{itemize}
\item Interpolation of ROBs:
 \[
        \mathbf V(\mathbf p_k^*) = \mathcal I_V\big(\{\mathbf V(\mathbf p^{(j)})\}_{j=1}^{N_\text{sam}^{\mathbf p}},\{\mathbf p^{(j)}\}_{j=1}^{N_\text{sam}^{\mathbf p}},\mathbf p_k^*\big);
        \]
\item Interpolation of PROMs:
  \[
        \mathbf L_r(\mathbf p_k^*) = \mathcal I_L\big(\{\mathbf L_r(\mathbf p^{(j)})\},\{\mathbf V(\mathbf p^{(j)})\}_{j=1}^{N_\text{sam}^{\mathbf p}},\{\mathbf p^{(j)}\}_{j=1}^{N_\text{sam}^{\mathbf p}},\mathbf p_k^*\big);
        \]
\item One-step prediction:
\[\text{Input:} \ \hat{\mathbf s}(t_k), \ \mathbf V(\mathbf p_k^*) \ \text{and}  \ \mathbf L_r(\mathbf p_k^*)\xrightarrow[\text{prediction}]{\text{DMD or Koopman augmented DMD}} \text{Output:} \ \hat{\mathbf s}(t_{k+1}) ;\]
\end{itemize}
\textbf{End}
\vspace{1mm}

\textbf{Output:} \(\mathcal S_\text{test}^\text{DRIPS} = \{\hat{\mathbf s}(t_0),\dots, \hat{\mathbf s}(t_{N_{T^*}})\}.\)
\end{algorithm}

\subsubsection{FML}
\label{fml}
An alternative methodology to discover the modified dynamics $\tilde{\mathbf f}^{\Delta t}(\mathbf s,\mathbf p)$ is Flow Map Learning~\cite{churchill2023flow,qin2021deep}. Given the training dataset~\eqref{eq:train_data}, FML seeks a nonlinear operator $\hat{\mathbf N}_\Theta(\mathbf s,\mathbf p): \Omega_{\mathbf s}\times \Omega_{\mathbf p}\to \Omega_{\mathbf s}$ that best approximates the evolution map 
\begin{equation}
\mathbf s^{(m,j)}_\text{out}\approx \widehat{\mathbf N}_\Theta(\mathbf s^{(m,j)}_\text{in},\mathbf p^{(j)}) \quad \text{ for all }\quad m = 0,\cdots, N_\text{sam}^{\mathbf s}, \quad j = 1,\cdots, N_\text{sam}^\mathbf p.
\end{equation}
where $\Theta$  are the network parameters that need to be trained. The neural network has a feed-forward architecture as illustrated in Figure~\ref{fig:fml-architecture}:
\begin{figure}[h!]
\includegraphics[width = 0.9\linewidth]{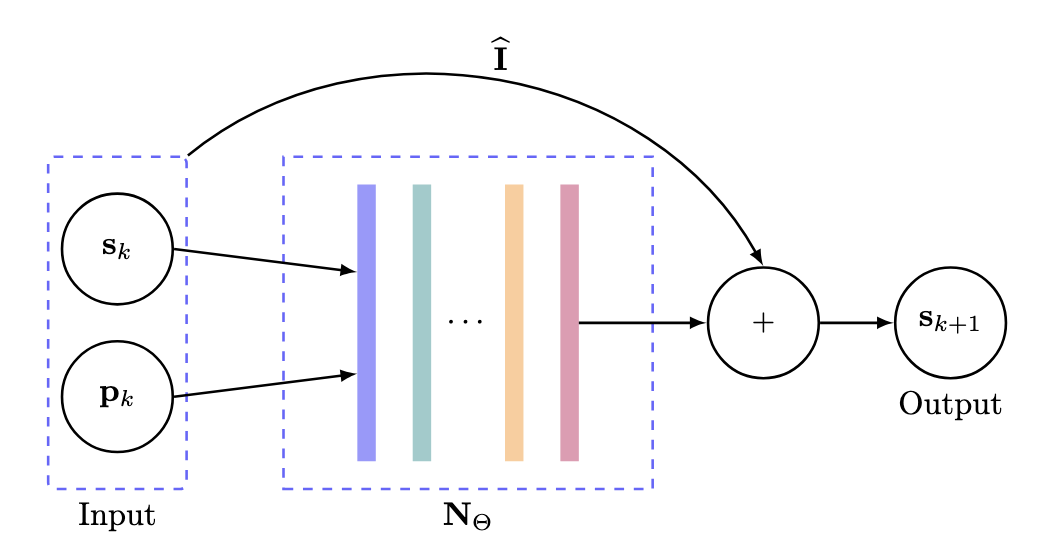}
\caption{Illustration of FML neural network architecture (adapted from~\cite{qin2021data}).}
\label{fig:fml-architecture}
\end{figure}

The input is multiplied by $\widehat {\mathbf I}$ and then reintroduced before the final output. The operator $\widehat {\mathbf I} $ is a matrix of size $N_s\times (N_s+N_\text{par})$. It takes the form
\begin{equation}
    \widehat {\mathbf I}  = [\mathbf I_{N_s},\mathbf 0],
\end{equation}
where $\mathbf I_{N_s}$ is identity matrix of size $N_s\times N_s$ and $\mathbf 0$ is a zero matrix of size $N_s\times N_\text{par}$. Therefore, the network effectively defines a mapping $\widehat{\mathbf N}_\Theta = (\widehat{\mathbf I} + \mathbf N_\Theta)$. Let us assume we have $M\geq 1$ hidden layers (the colored layers in Figure~\ref{fig:fml-architecture}), each of which contains $d\geq 1$ nodes. The DNN operator $\mathbf N_\Theta$ can be written as the following composition of affine and nonlinear transformations:
\begin{equation}
    \mathbf N_\Theta = \mathbf W_M \circ (\sigma_M \circ \mathbf W_{M-1})\circ \cdots \circ (\sigma_1\circ \mathbf W_0),
\end{equation}
where $\mathbf W_i$ is the weight matrix containing the weights connecting the $i^\text{th}$ and $(i+1)^\text{th}$ layers and the biases in the $(i+1)^\text{th}$ layer, $\sigma_i$ is the (nonlinear) component-wise activation function of the $i^\text{th}$ layers, and the $0^\text{th}$ layer is the input layer. At the output layer, the $(M+1)^\text{th}$ layer, the linear activation function ($\sigma(x) = x$) is used.

Let us define a shorthand notation 
\begin{equation}\label{eq:W}
    \mathbf W[0:m] = [\mathbf W_0,\cdots, \mathbf W_m],\quad 0\leq m\leq M.
\end{equation}
The hyperparameters $\Theta$ then refer to the collection of all the weight matrices, i.e., $\Theta = \mathbf W[0:M]$. The network training is conducted by minimizing the mean squared loss between the network output $\widehat{\mathbf N}_\Theta(\mathbf s^{(m,j)}_\text{in},\mathbf p^{(j)})$ and the data $\mathbf s^{(m,j)}_\text{out}$, i.e., 
\begin{equation}\label{eq:NN_par}
    \Theta^* = \arg\min_\Theta \frac{1}{N_\text{sam}^{\mathbf s}N_\text{sam}^{\mathbf p}}\sum_{m = 1}^{N_\text{sam}^{\mathbf s}}\sum_{j = 1}^{N_\text{sam}^{\mathbf p}}\|\widehat{\mathbf N}_\Theta(\mathbf s^{(m,j)}_\text{in},\mathbf p^{(j)})-\mathbf s^{(m,j)}_\text{out}\|^2.
\end{equation}
On satisfactory training of the network parameter using~\eqref{eq:NN_par}, we obtain a trained network model for the unknown modified system~\eqref{eq:modified-HFM-dis},
\begin{equation}\label{eq:fml-test}
    \mathcal S_\text{test}^\text{FML} = \{\hat{\mathbf s}(0), \cdots, \hat{\mathbf s}(t_{N_{T^*}})\},
\end{equation}
where the state is advanced recursively by the learned map $\widehat{\mathbf N}_{\Theta^*}$ according to
\begin{equation}\label{eq:FML-pred}
\left\{
\begin{aligned}
    &\hat{\mathbf s}(0) = \mathbf s(0),\\
    &\hat{\mathbf s}(t_{k+1}) = \widehat{\mathbf N}_{\Theta^*}(\hat{\mathbf s}(t_k),\mathbf p_k^*)\quad \text{for} \quad k = 0,\cdots, N_{T^*}-1.
    \end{aligned}
    \right.
\end{equation}

\begin{remark}
   The DRIPS predicting model~\eqref{eq:DRIPS-pred} and the FML predicting model~\eqref{eq:FML-pred} are approximations to the one-step evolution~\eqref{eq:modified-HFM-dis} of the modified system~\eqref{eq:modified}, which in turn is an approximation of the original unknown dynamical system~\eqref{eq:general}. Therefore, \eqref{eq:DRIPS-pred} and \eqref{eq:FML-pred} generate approximations to the solution of the unknown system~\eqref{eq:general} at the discrete time instance $t_k$ in~\eqref{eq:test_time}.
\end{remark}

\begin{remark}
The DRIPS framework can be regarded as a special case of the FML model with an explicitly enforced linear structure. Owing to this structure, DRIPS typically requires significantly fewer training samples than FML as reported in~\cite{lu2024data} and~\cite{qin2021data}. Moreover, DRIPS allows for the easier incorporation of physical knowledge, for instance, through the choice of the lifting map. In contrast, FML provides a more flexible and general framework capable of capturing stronger nonlinearities and accommodating nonuniform time stepping in the evolution of the learned dynamics, as demonstrated in~\cite{qin2021data}. Consequently, DRIPS offers efficiency and interpretability, while FML achieves greater expressivity and accuracy at higher data demands.
\end{remark}

\section{FML Model Correction}\label{sec:correction}
An extension to the FML methodology was introduced in \cite{tatsuoka2025deep} for correcting a prior flow-map model with a scarce high-fidelity dataset via transfer learning; the resulting corrected flow-map model can conduct improved predictions of the true underlying dynamics. This section briefly reviews the method, and further details can be found in \cite{tatsuoka2025deep}. We begin by assuming that we have in our possession an initial \textit{low-fidelity} dynamical system model, where the low-fidelity system may be imperfect due to simplified physics, linear approximations to nonlinear dynamics, and/or incorrect parameterizations. This low-fidelity model shall be referred to as the ``prior model". In this work, we consider the prior model~\eqref{eq:prior} to be an \textit{incorrect parameterization} of the true underlying system~\eqref{eq:general}.

The first step is to construct an FML model for the modified representation of the prior system~\eqref{eq:prior}:
\begin{equation}\label{eq:prior}
\left\{
\begin{aligned}
&
\frac{d\tilde {\mathbf{s}}(t)}{dt} = \mathbf{f}_\text{prior}(\tilde{\mathbf{s}}(t), \tilde{\mathbf{c}}(t;\mathbf p)), \quad t \in (0, T],\\
&\tilde{\mathbf{s}}(0) = \mathbf s_0,
\end{aligned}
\right.
\end{equation}
where the local parameterization of the control input is performed in the same manner as in~\eqref{eq:HFM-dis}–\eqref{eq:modified}.

% We again recall Lemma 3.1 in \cite{qin2021data} and aim to approximate  
% To construct the DNN prior model, we repeatedly execute the prior model \eqref{prior} to generate a large number of data pairs separated by time lag $\Delta t$; 
Following equations Eq.~\eqref{eq:flow_map_true}-\eqref{eq:training_data} in Section~\ref{sec:dataset}, we consider trajectories generated by our prior model as well as the local parameterization of the model input to build the prior training dataset $\mathcal{{S}}^{\text{prior}}_\text{train}$ of size $J^{LF}$. Since we have in our possession the prior model, $J^{LF}$ can be very large.
%\begin{equation}\label{eq:training_data_LF}
%\mathcal{{S}}^{\text{prior}}_\text{train} = \bigcup\limits_{m,j} \mathcal{\tilde{S}}_\text{train}^{(m,j)},  \qquad m = 0,\dots N_\text{sam}^{\tilde{\mathbf{s}}}, \qquad j = 1,\dots, N_\text{sam}^{\mathbf p},
%\end{equation}
%where 
%\begin{equation}\label{eq:training_data_LF_sub}
%\mathcal{\tilde{S}}_\text{train}^{(m,j)} = \{\{\mathbf{\tilde{s}}^{(m,j)}_\text{in},\mathbf p^{(j)}\} \; , \; \{\mathbf{\tilde{s}}^{(m,j)}_\text{out}\}\}.
%\end{equation}
%We denote the total size of the low-fidelity data set as $J^{LF} = N_\text{sam}^{\tilde{\mathbf{s}}} \times N_\text{sam}^{\mathbf p}$.
Using the low-fidelity dataset $\mathcal{{S}}^{\text{prior}}_\text{train}$, we then construct a FML model following Section~\ref{fml}. Let $\widetilde{\mathbf{N}}_{\Theta^*_\text{prior}}:\mathbb{R}^{N_s + N_\text{par}} \rightarrow \mathbb{R}^{N_s}$ be the trained DNN approximation where $\Theta^*_\text{prior}$ are the trained network parameters obtained from FML on the prior training dataset $\mathcal{{S}}^{\text{prior}}_\text{train}$. It follows that the trained model $\widetilde{\mathbf{N}}_{\Theta^*_\text{prior}}$ acts as a one-step flow map for the prior system~\eqref{eq:prior}. 
%\begin{equation}\label{DNN_prior}
%\mathbf{\tilde{s}}^{(m,j)}_\text{out}\approx \widetilde{\mathbf N}_{\Theta^*}(\mathbf{\tilde{s}}^{(m,j)}_\text{in},\mathbf p^{(j)}) \quad \text{ for all }\quad m = 0,\cdots, N_\text{sam}^{\mathbf{\tilde{s}}}, \quad j = 1,\cdots, N_\text{sam}^\mathbf p.
%\end{equation}

\begin{comment}
Recall the shorthanded notation introduced in Eq.~\eqref{eq:W}
$$
\mathbf{W}_{[0:m]} = [\mathbf{W}_0, \dots, \mathbf{W}_m], \qquad 0\leq m\leq M.
$$
where the hyperparameters $\Theta$ refer to the collection of all the weight matrices, i.e.,
$\Theta = \mathbf{W}_{[0:M]}$. 
The parameter optimization problem in (30) can be written equivalently as

\begin{equation}\label{eq:NN_par_LF}
    \widetilde{\mathbf{W}}^*_{[0:M]} = \arg\min_{{\mathbf{W}}_{[0:M]}} \frac{1}{N_\text{sam}^{\mathbf {\tilde{s}}}N_\text{sam}^{\mathbf p}}\sum_{m = 1}^{N_\text{sam}^{\mathbf {\tilde{s}}}}\sum_{j = 1}^{N_\text{sam}^{\mathbf p}}\|\widetilde{\mathbf N}(\mathbf{\tilde{s}}^{(m,j)}_\text{in},\mathbf p^{(j)}; \mathbf{W}_{[0:M]})- \mathbf{\tilde{s}}^{(m,j)}_\text{out}\|^2.
\end{equation}

\end{comment}

We now describe the step for obtaining the high-fidelity flow-map model: the basic premise is that the trained DNN prior model $\widetilde{\mathbf{N}}_{\Theta^*_\text{prior}}$, which is an accurate representation of the prior model \eqref{eq:prior}, can capture the ``bulk" behavior of the dynamics of the unknown system \eqref{eq:modified}. To further correct the DNN prior model, we employ transfer learning (TL) technique, with the help of the scarce high-fidelity data set $\mathcal S_\text{train}$ in~\eqref{eq:train_data}, obtained from the true dynamics~\eqref{eq:general}; the total number of high-fidelity training data pairs as $J^{HF} = N_\text{sam}^{{\mathbf{s}}} \times N_\text{sam}^{\mathbf p}$. The principle of transfer learning (TL) is based on the widely accepted notion that the early layers of a DNN extract more general features of a dataset, while later layers contain higher-level features.
Following this, we ``freeze" some layers in the DNN prior model $\widetilde{\mathbf{N}}_{\Theta^*_\text{prior}}$. Specifically, we fix the weights and biases in some of the layers of the trained DNN prior model by making them un-modifiable. We then use the high-fidelity data set to retrain the parameters in the last few layers. For example, in Figure~\ref{fig:fml-architecture}, the blue and green layers are frozen, while the yellow and red layers are retrained.

Let $0\leq \ell\leq M$ be a number separating the layers in the trained DNN prior model $\widetilde{\mathbf{N}}_{\Theta^*_\text{prior}}$ into two groups: the first $\ell$ layers from the input layer ($0^{th}$ layer) to the $(\ell-1)^{th}$ layer, and the second group from the $\ell^{th}$ layer to the output layer ($M^{th}$ layer).
Using the notation defined in Eq.\eqref{eq:W}, the hyperparameters can be 
separated into the following two groups correspondingly, 
\begin{equation} \label{W_split}
    \Theta^*_\text{prior} = \mathbf{W}^*_\text{prior}[0:M] = \left[\mathbf{W}^*_\text{prior}[0:\ell-1], \mathbf{W}^*_\text{prior}[\ell:M]\right].
\end{equation}

We fix the first group of parameters to be at the values trained in the DNN prior model $\widetilde{\mathbf{N}}_{\Theta^*_\text{prior}}$, i.e. $\mathbf{W}[0:\ell-1] = \mathbf{W}^*_\text{prior}[0:\ell-1]$, and re-train the second group of parameters by minimizing the mean square error of the model~\eqref{eq:NN_par} over the high-fidelity data set \eqref{eq:train_data}. The optimization problem then becomes 
{\small
\begin{equation}\label{eq:NN_par_HF}
\begin{aligned}
    \mathbf{W}^*_\text{pos}[\ell:M]
    = \arg\min_{\mathbf{W}_{[\ell:M]}} 
    \frac{1}{N_\text{sam}^{\mathbf{s}} N_\text{sam}^{\mathbf{p}}}
    \sum_{m = 1}^{N_\text{sam}^{\mathbf{s}}}
    \sum_{j = 1}^{N_\text{sam}^{\mathbf{p}}}
    \left\|
        \widetilde{\mathbf{N}}\!\left(
            \mathbf{s}^{(m,j)}_\text{in}, \mathbf{p}^{(j)};
            \widetilde{\mathbf{W}}^*_\text{prior}[0:\ell-1], \mathbf{W}[\ell:M]
        \right)
        - \mathbf{s}^{(m,j)}_\text{out}
    \right\|^2.
\end{aligned}
\end{equation}
}

Once training is completed, we obtain a DNN whose hyperparameters are
\begin{equation} \label{W_final}
\Theta^* = \mathbf{W}^*[0:M] =\left[ \mathbf{W}^*_\text{prior}[0:\ell-1], \mathbf{W}^*_\text{pos}{[\ell:M]}\right],
\end{equation}
where the first group is from~\eqref{W_split} and the second group is from \eqref{eq:NN_par_HF} separately. Finally, we obtain our posterior DNN model $\hat{\mathbf N}_{\Theta^*}$, which allows for the same prediction as in~\eqref{eq:FML-pred}.

%operator $\mathbf{N}$ using these parameters, i.e.,
%\begin{equation} \label{DNN-post}
 %   \mathbf{N} \triangleq \widetilde{\mathbf{N}}\left(\cdot; \mathbf{W}^*_{[0:M]}\right).
%\end{equation}
 
\section{Numerical Examples}\label{sec:num-ex}
We evaluate the proposed frameworks on the vehicle dynamics introduced in Section~\ref{sec:dynamicsOverview}, considering multiple levels of model complexity and data fidelity. 

\medskip
\noindent\textbf{Case I: Perfectly represented dynamics.}  
We first assume that the simplified approximation models perfectly represent the true dynamics, i.e., $\mathbf f_\text{true} = \mathbf f_\text{prior}$. Synthetic datasets are generated from the known form  $\mathbf f_\text{prior}$ under various prescribed control inputs to validate the learning methodologies. The training data are produced by numerically integrating the known dynamical systems~\eqref{eq:unicyle}–\eqref{eq:full-bicyle} using a high-resolution solver implemented in JAX (e.g., \texttt{Diffrax's Tsit5}, an explicit fifth-order Runge–Kutta scheme with an embedded fourth-order adaptive step sizing method). Both DRIPS and FML are then tested for their prediction capabilities under unseen control inputs across models of increasing complexity, as discussed in Sections~\ref{sec:unicycle-drips},~\ref{sec:sim-bicyle-drips}, and~\ref{sec:full-bicyle-fml}, respectively.

\medskip
\noindent\textbf{Case II: Parameterized model discrepancy.}  
Next, we assume that the true dynamics share the same physics-based formulation as the prior model, but differ in parameter values $\gamma$ (e.g., scaling coefficients $b_u$ and $b_\delta$). In this case,
\[
\mathbf f_\text{true} = \mathbf f(\mathbf s(t), \mathbf c(t); \gamma_\text{true}), \qquad
\mathbf f_\text{prior} = \mathbf f(\mathbf s(t), \mathbf c(t); \gamma_\text{prior}), \qquad
\gamma_\text{true} \neq \gamma_\text{prior}.
\]
A large dataset $\mathcal S_\text{train}^\text{prior}$ and a small dataset $\mathcal S_\text{train}$ are generated from $\mathbf f_\text{prior}$ and $\mathbf f_\text{true}$, respectively, using the same numerical scheme as before. The model-correction methodology introduced in Section~\ref{sec:correction} is validated in Sections~\ref{sec:correction-sim} and~\ref{sec:correction-full}.

\medskip
\noindent\textbf{Case III: Experimentally observed dynamics.}  
Finally, in Section~\ref{sec:exp-data}, we take laboratory measurement data as $\mathcal S_\text{train}$, reflecting an entirely unknown dynamical process $\mathbf f$ that carries substantial and intricate model-form uncertainty when compared with $\mathbf f_\text{prior}$. The dataset $\mathcal S_\text{train}^\text{prior}$ is generated by numerical simulation of $\mathbf f_\text{prior}$, while a limited number of experimental observations $\mathcal S_\text{train}$ are used to correct the prior model.

\medskip
\noindent\textbf{Data generation and sampling.}  
We adopt local parameterization via interpolating polynomials over equally spaced points following~\cite{lu2024data, qin2021data}. The training dataset is organized in the form of~\eqref{eq:train_data}.  

For DRIPS, the parameter values are sampled on a Cartesian grid over the $N_\text{par}$-dimensional parameter space, with three points per dimension (two endpoints and one midpoint), resulting in $N_\text{sam}^\mathbf p = 3^{N_\text{par}}$, as in~\cite{lu2024data}. In contrast, FML draws parameter samples uniformly from $\Omega_\mathbf p$, typically generating datasets of size $10^5$–$10^6$, which is substantially larger than those used for DRIPS when $N_\text{par}$ is moderate~\cite{qin2021data}.

For each local subset $\mathcal S_\text{train}^{(m,j)}$, the initial state $\mathbf s^{(m)}_\text{in}$ is randomly sampled from $\Omega_\mathbf s$ using a uniform distribution, and the subsequent state $\mathbf s^{(m,j)}_\text{out}$ is obtained by solving the reference system with a time step $\Delta t$ and a control input parameterized as in~\eqref{eq:local_par}. Unless stated otherwise, all numerical examples use $\Delta t = $\SI{0.01}{\second}. The sampling domains $\Omega_{\mathbf s}$ and $\Omega_{\mathbf p}$ are determined from prior knowledge of the underlying dynamics to ensure that the target trajectories lie within $\Omega_{\mathbf s}$ for all $\mathbf p \in \Omega_{\mathbf p}$, satisfying the assumptions required in the theoretical analysis of~\cite{qin2021data}. The total number of training samples, as well as the specific ranges of $\Omega_{\mathbf s}$ and $\Omega_{\mathbf p}$, are reported for each example below.

\medskip
\noindent\textbf{Model evaluation.}  
Once the surrogate model is trained, predictions are obtained iteratively using~\eqref{eq:DRIPS-pred} or~\eqref{eq:FML-pred} with new initial conditions and control inputs. The resulting trajectories are compared against reference solutions computed from the exact system using the same control inputs. For model-correction tasks, predictions from the corrected model $\hat{\mathbf N}_{\Theta^*}$, with $\Theta^*$ defined in~\eqref{W_final}, are compared against reference or experimental data under identical control conditions. The modified relative error is computed to evaluate the performance for each state variable as 
\begin{equation}\label{eq:modi-l2-err} 
    \mathcal E_i(t_k) = \frac{|s_i(t_k)-{\hat s}_i(t_k)|}{|s_i(t_k)|+\epsilon},\quad i  =1,\cdots, N_s,\quad k = 1,\cdots,N_{T^*},
\end{equation}
where $s_i(t_k)$ denotes the $i^\text{th}$ state in $\mathbf s(t_k)$ of $\mathcal S_\text{test}$ in~\eqref{eq:data_test} and $\hat{s}_i(t_k)$ denotes the $i^\text{th}$ state in $\hat{\mathbf s}(t_k)$ of DRIPS prediction or FML prediction/correction. A small positive number $\epsilon$ is added in the denominator to prevent a blow-up near $|s_i(t_k)|=0$.

\subsection{Learning Unicycle Dynamics via DRIPS}\label{sec:unicycle-drips}
This subsection investigates Case I, where we assume the prior unicycle model~\eqref{eq:unicyle} perfectly represents the true dynamics, i.e., $\mathbf f_\text{true} = \mathbf f_\text{prior}$. Due to the simplicity of the unicycle system, we employ the DRIPS framework, which offers higher data efficiency than FML. Here, we validate its capability as a data-driven learning tool to predict system trajectories under previously unseen control inputs. 

The time-dependent inputs $v_x(t)$ and $\omega(t)$ in the unicycle dynamics~\eqref{eq:unicyle} are locally parameterized with second-degree polynomials. As a result, the local parameter set~\eqref{eq:local_par} $\mathbf p_k\in \mathbb R^{N_\text{par}}$ with $N_\text{par} = 3+3 = 6$. The training data are generated from $N_\text{sam}^{\mathbf p} = 2^6$ parameter points on the Cartesian grid of the parameter space $\Omega_{\mathbf p} = [-1,1]^6$. For each parameter point $\mathbf p^{(j)}$, $N_\text{sam}^{\mathbf s} = 6$ pairs of data in the form of~\eqref{eq:train_data_sub} are generated with an initial state $\mathbf s^{(m)}_\text{in}$ randomly drawn from the state variable space $\Omega_{\mathbf s} = [-0.6,8]\times[0,2]\times [-0.6,2\pi]$. The total number of training data pairs is 384. 

We choose the lifting map $\mathbf g(t) = [x(t), y(t), \yaw(t), \cos(\yaw(t)), \sin(\yaw(t))]^\top$ so that the lifted dynamics take the linear form
\begin{equation}\label{eq:obsUniDynamics}
    \frac{d \mathbf g(t)}{dt}=\begin{bmatrix}0 & 0 & 0 & v_x(t) & 0 \\ 0 & 0 & 0 & 0 & v_x(t) \\ 0 & 0 & 0 & 0 & 0 \\ 0 & 0 & 0 & 0 & -\omega(t) \\ 0 & 0 & 0 & -\omega(t) & 0\end{bmatrix}\mathbf{g}(t)+\begin{bmatrix} 0 \\ 0 \\ \omega(t) \\ 0 \\ 0 \end{bmatrix}.
\end{equation}

\begin{figure}
    \centering
    \includegraphics[width=\linewidth]{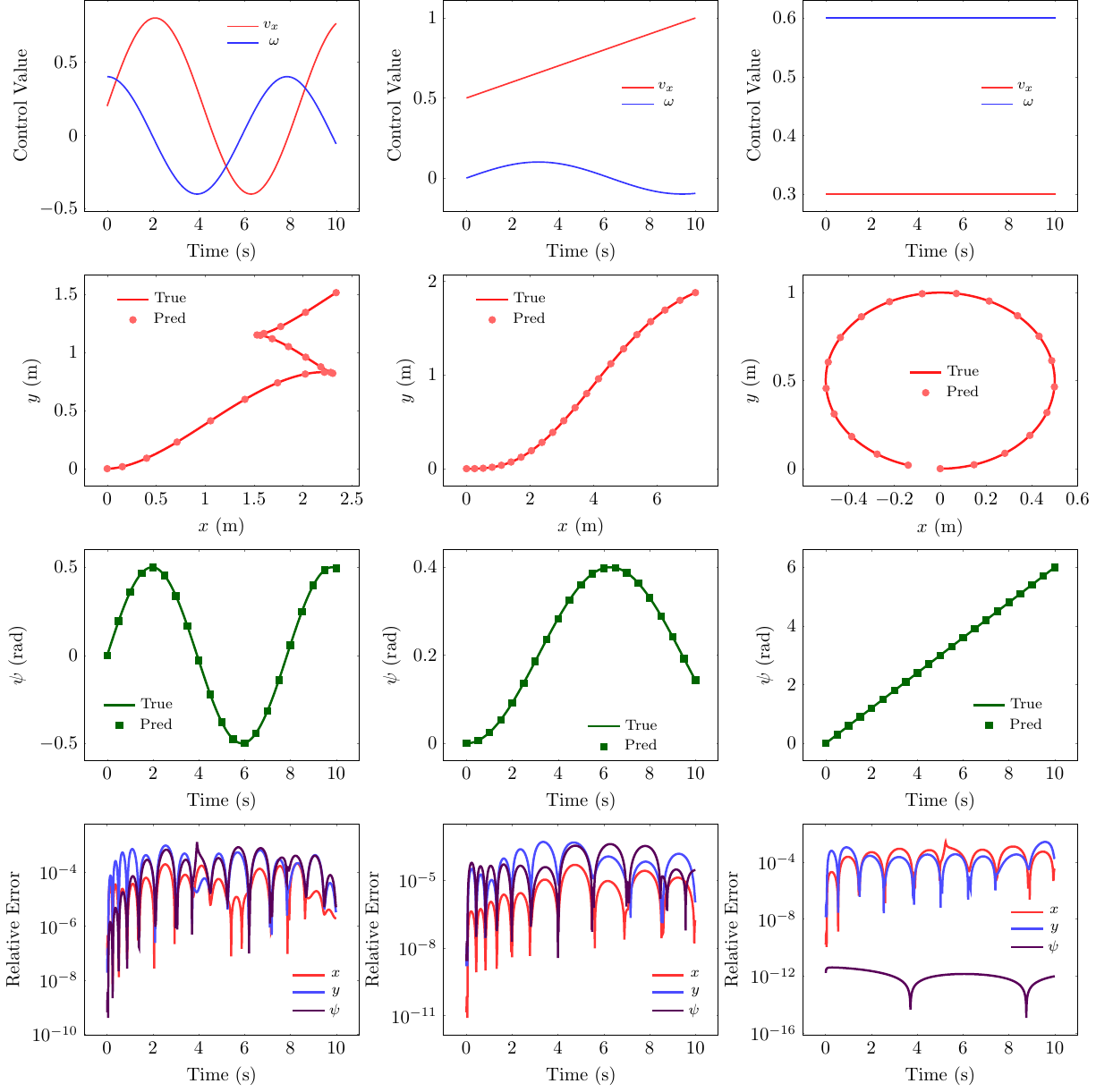}
    \caption{DRIPS prediction for unicycle dynamics~\eqref{eq:unicyle} with testing control inputs: Left: (a) sinusoidal turning~\eqref{eq:control-sin}; Middle: (b) linear speedup with oscillating steering~\eqref{eq:control-lin}; Right: (c) circular pattern at constant speed~\eqref{eq:control-circ}. Top: true yaw angle vs. DRIPS prediction yaw angle and control inputs over time; Middle: true trajectory vs. DRIPS predicted trajectory; Bottom: prediction error of the DRIPS surrogate model, as defined in~\eqref{eq:modi-l2-err}.}
    \label{Unicycle}
\end{figure}

Once the DRIPS surrogate is trained, we use it to predict the system behavior over the time horizon $T^* = [0,10]$s with the initial conditions: $x(0)=0,y(0)=0,\psi(0)=0$. Three new control input profiles, unseen during training, are considered:
\begin{align}
    &\text{(a) Sinusoidal turning}:
    \begin{cases}\label{eq:control-sin}
        v_x(t)=0.2+0.6\sin(0.75t), \\
        \omega(t)=0.4\cos(0.8t),
    \end{cases} \\
    &\text{(b) Linear speedup with oscillating steering}:
    \begin{cases}\label{eq:control-lin}
        v_x(t)=0.5+0.05t, \\
        \omega(t)=0.1\sin(0.5t),
    \end{cases} \\
    &\text{(c) Circular pattern at constant speed}:
    \begin{cases}\label{eq:control-circ}
        v_x(t)=1.0, \\
        \omega(t)=0.2.
    \end{cases}
\end{align}
In Figure~\ref{Unicycle}, the DRIPS predictions are visually indistinguishable from the reference trajectories. The modified relative error defined in~\eqref{eq:modi-l2-err} for each state variable remains in the range of $10^{-15}$-$10^{-2}$ across all test scenarios, demonstrating the high accuracy and generalization capability of the DRIPS surrogate model.

\begin{comment}
\begin{figure}
    \centering
    \includesvg[width=0.3\linewidth]{figures/DRIPS Figures/UnicycleSet1.svg}
     \includesvg[width=0.3\linewidth]{figures/DRIPS Figures/UnicycleSet2.svg}
     \includesvg[width=0.3\linewidth]{figures/DRIPS Figures/UnicycleSet3.svg}
    \includesvg[width=0.3\linewidth]{figures/DRIPS Figures/UnicycleSet1_Traj.svg}
        \includesvg[width=0.3\linewidth]{figures/DRIPS Figures/UnicycleSet2_Traj.svg}
     \includesvg[width=0.3\linewidth]{figures/DRIPS Figures/UnicycleSet3_TrajDRIPS.svg}
     \includesvg[width=0.3\linewidth]{figures/DRIPS Figures/Unicycle_ErrSet1.svg}
        \includesvg[width=0.3\linewidth]{figures/DRIPS Figures/Unicycle_ErrSet2.svg}
     \includesvg[width=0.3\linewidth]{figures/DRIPS Figures/Unicycle_ErrSet3.svg}
    \caption{DRIPS prediction for unicycle dynamics~\eqref{eq:unicyle} with testing control inputs: Left: (a). sinusoidal turning~\eqref{eq:control-sin}; Middle: (b). linear speedup with oscillating steering~\eqref{eq:control-lin}; Right: (c). circular pattern at constant speed~\eqref{eq:control-circ}. Top: true yaw angle vs. DRIPS prediction yaw angle and control inputs over time; Middle: true trajectory vs. DRIPS prediction trajectory; Bottom: prediction error (as defined in~\eqref{eq:modi-l2-err}) of the DRIPS surrogate model.}
    \label{Unicycle}
\end{figure}
\end{comment}

\subsection{Simplified Planar Bicycle Model}
Now consider \eqref{eq:simp-bicyle} with initial conditions: $x(0)=0,y(0)=0,v_x(0)=0,\psi(0)=0$, we test on the following new control input settings:
\begin{align}
    &\text{(a) Coupled oscillations}:
    \begin{cases}\label{eq:control-coup-os}
        u(t)=0.1\sin(0.4t), \\
        \delta(t)=-0.3\cos(0.6t),
    \end{cases} \\
    &\text{(b) Slow ramp throttle with constant steering}:
    \begin{cases}\label{eq:control-slow-ramp}
        u(t)=0.05t\exp(-0.3t), \\
        \delta(t)=0.2,
    \end{cases} \\
    &\text{(c) Pulse acceleration and steering}:
    \begin{cases}\label{eq:control-pulse}
        u(t)=0.1\sin^2(0.5t),\\
        \delta(t)=0.25\cos^2(0.25t).
    \end{cases}
\end{align}

\subsubsection{Learning Simplified Planar Bicycle Model via DRIPS}\label{sec:sim-bicyle-drips}
We first repeat the evaluation of Case I, assuming the prior simplified planar bicycle model~\eqref{eq:simp-bicyle} perfectly represents the true dynamics with the true parameter values $\gamma_\text{true} = \gamma_\text{prior} = [b_u,b_\delta, L]^\top = [4.55,0.4601, 0.255]^\top$.

The time-dependent control inputs $u(t)$ and $\delta(t)$ in the dynamics~\eqref{eq:simp-bicyle} are locally parameterized using second-degree polynomials, resulting in the local parameter vector $\mathbf p_k \in \mathbb R^{N_\text{par}}$ with $N_\text{par} = 3 + 3 = 6$. Given the moderate complexity of the simplified bicycle model, DRIPS is again adopted as the learning framework due to its superior data efficiency compared to FML. The training data are generated from $N_\text{sam}^{\mathbf p} = 2^6$ parameter points distributed on a Cartesian grid within the parameter space $\Omega_{\mathbf p} = [-0.3,0.3]^6$. For each parameter point $\mathbf p^{(j)}$, $N_\text{sam}^{\mathbf s} = 13$ data pairs in the form of~\eqref{eq:train_data_sub} are obtained, with initial states $\mathbf s^{(m)}_\text{in}$ uniformly sampled from $\Omega_{\mathbf s} = [-4,10]\times[-1,8]\times[0,2.5]\times[-0.5,4]$. The total number of training pairs is 832. 

\(\mathbf{g} = [x, y, v_x,\psi, \cos(\psi), \sin(\psi), v_x\cos(\psi), v_x\sin(\psi), v_x^2, \cos^2(\psi), \sin^2(\psi), \sin(\psi)\cos(\psi)]^\top \) is chosen as the lifted observable. The resulting lifted linear dynamics are given in the Appendix. 

In Figure~\ref{Simp-Bicycle}, the DRIPS predictions are visually indistinguishable from the reference trajectories. The modified relative error defined in~\eqref{eq:modi-l2-err} for each state variable remains in the range of $10^{-8}$-$10^{-3}$ across all test scenarios, demonstrating the high accuracy and generalization capability of the DRIPS surrogate model.

% \begin{figure}
%     \includegraphics[width=\linewidth]{CSV/Figure-Examples.pdf}
% \end{figure}

\begin{figure}
    \centering
    \includegraphics[width=\linewidth]{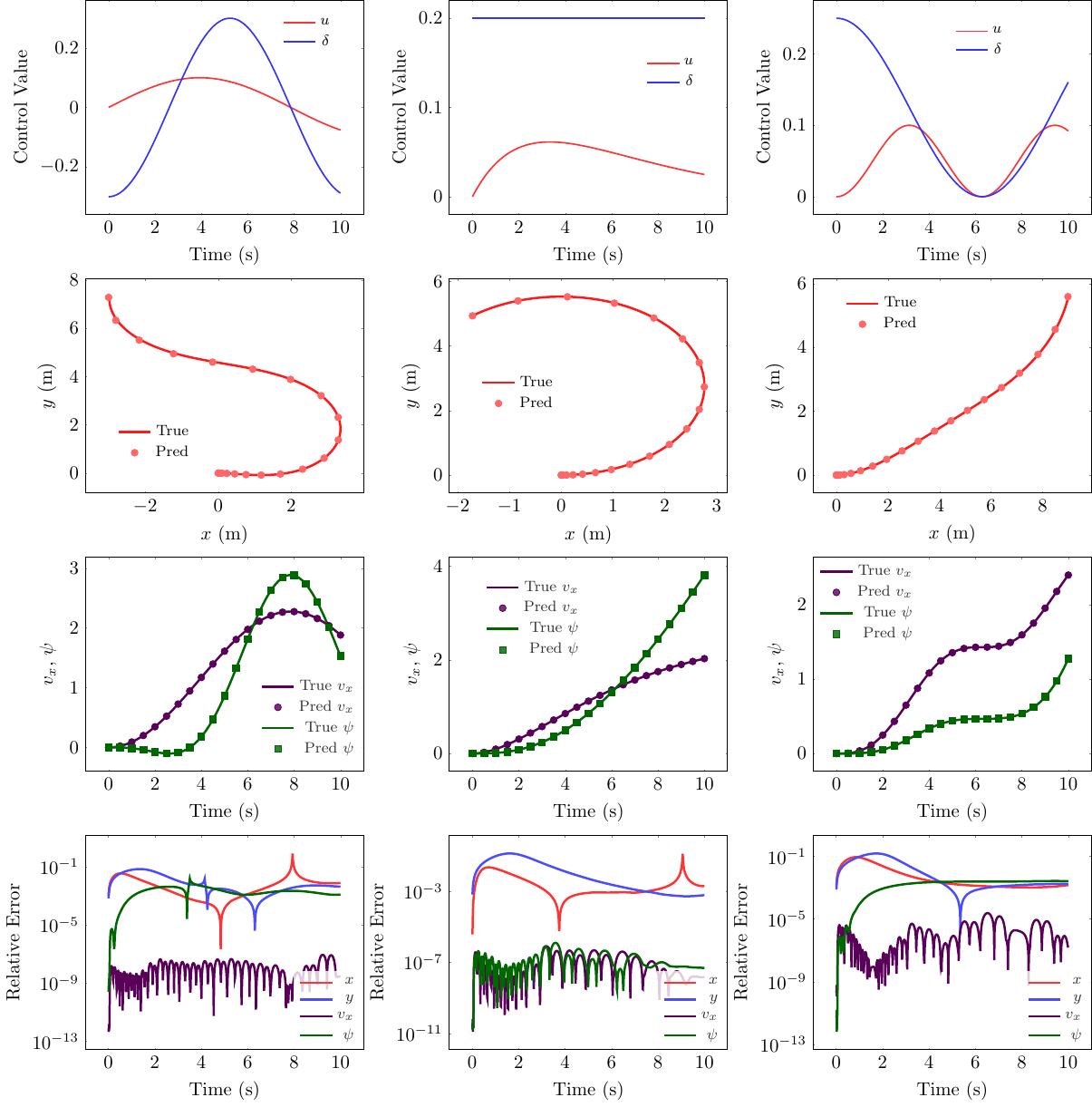}   
    \caption{DRIPS prediction for simplified planar bicycle model~\eqref{eq:simp-bicyle} with testing control inputs: Left: (a) coupled oscillations~\eqref{eq:control-coup-os}; Middle: (b) slow ramp throttle with constant steering~\eqref{eq:control-slow-ramp}; Right: (c) Pulse acceleration and steering ~\eqref{eq:control-pulse}. Row 1: control inputs over time; Row 2: true trajectory vs. DRIPS prediction trajectory; Row 3: true forward velocity and yaw angle vs. DRIPS prediction forward velocity and yaw angle over time; Row 4: prediction error of the DRIPS surrogate model, as defined in~\eqref{eq:modi-l2-err}.}
    \label{Simp-Bicycle}
\end{figure}

\subsubsection{Correcting Model Error in Simplified Planar Bicycle Model via FML}\label{sec:correction-sim}
This subsection investigates Case II, where we assume the prior model~\eqref{eq:simp-bicyle} describes the true dynamics formulation correctly but misparameterizes the system, i.e., $\gamma^\text{true} = [b_u^\text{true},b_\delta^\text{true}, L^\text{true}]^\top = [1,\pi/6, 0.3]^\top$ and $\gamma_\text{prior} = [b_u^\text{prior},b_\delta^\text{prior}, L^\text{prior}]^\top = [1.5,\pi/5, 0.5]^\top$. We wish to correct the prior flow map model using a scarce high-fidelity dataset.  

To train the prior flow map, the initial condition domain is taken to be \(\Omega_\mathbf{\mathbf{s}} = [-15,15] \times [-15,15] \times [-5,5] \times [-5,5]\). To approximate the one-step evolution operator, quadratic polynomials are used for local parameterization in the FML modeling of both $u(t)$ and $\delta(t)$, where we employ Legendre orthogonal polynomials; specifically, we employ tensor Legendre orthogonal
polynomials in total degree space, taking the coefficient space 
to be \(\Omega_{\mathbf p}= [-0.5,0.5]^6\). 
%The constant time-lag is taken to be $\Delta = 0.1$. 
The DNN is taken to be 3 hidden layers, each containing 80 nodes. The network training is conducted using $J_{LF} = 50,000$ data trajectories randomly from $\Omega_\mathbf{s} \times \Omega_{\mathbf p}$ to build the low-fidelity dataset $\mathcal S_\text{prior}^\text{train}$. The model is trained for $10^4$ epochs using the ADAM optimizer and a learning rate of $10^{-3}$, and we obtain the prior model $\widetilde{\mathbf{N}}_{\Theta^*_\text{prior}}$.

\begin{figure}
    \centering
    \includegraphics[width=\linewidth]{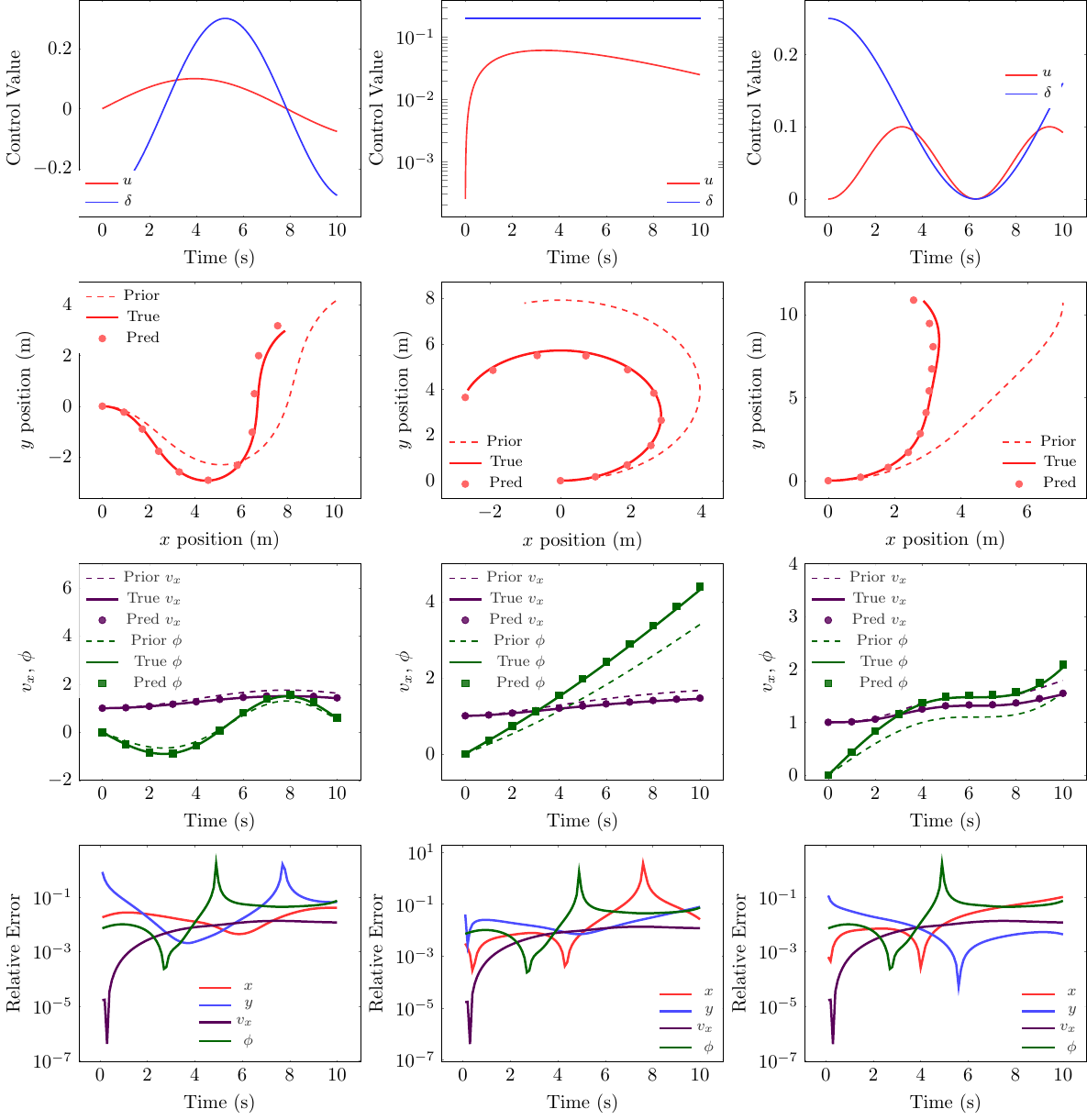}   
    \caption{FML model correction for simplified planar bicycle model~\eqref{eq:simp-bicyle} with testing control inputs: Left: (a) coupled oscillations~\eqref{eq:control-coup-os}; Middle: (b) slow ramp throttle with constant steering~\eqref{eq:control-slow-ramp}; Right: (c) Pulse acceleration and steering ~\eqref{eq:control-pulse}. Row 1: control inputs over time; Row 2: true/prior/FML prediction trajectory; Row 3: true/prior/FML prediction forward velocity and yaw angle over time; Row 4: prediction error  of the FML model, as defined in~\eqref{eq:modi-l2-err}.}
    \label{Simp-Bicycle-fml}
\end{figure}

For model correction, we consider a scarce high-fidelity data set of only $J_{HF} = 500$ samples using the correct parameterization $\gamma_\text{true} = [b_u^\text{true},b_\delta^\text{true}, L^\text{true}]^\top = [1,\pi/6, 0.3]^\top$. We fix the first layer and conduct transfer learning on the final two layers, using the loss function defined in~\eqref{eq:NN_par_HF}; we train for at most 5000 epochs with early stopping (patience of 100) to obtain the corrected flow map $\widetilde{\mathbf{N}}_{\Theta^*}$. In Figure~\ref{Simp-Bicycle-fml}, for the test initial condition $x(0)=0,y(0)=0,v_x(0)=0,\psi(0)=0$, we observe good agreement and the ability of the method to correct the dynamics due to the initial misparametrization.

\begin{comment}
\begin{figure}
\includegraphics[width=\linewidth]{CSV/slip-free-planar-model-model-correction/simp-bike-input1.pdf}

    \caption{FML model  correction for simplified planar bicycle model~\eqref{eq:simp-bicyle} with coupled oscillations~\eqref{eq:control-coup-os} test control input; Top: true/prior/FML prediction trajectory and relative error for $x$ and $y$ over time; Bottom: true/prior/FML prediction and relative error for forward velocity and yaw angle  over time.}
    \label{Simp-Bicycle-correct1}
\end{figure}

\begin{figure}
  \includegraphics[width=\linewidth]{CSV/slip-free-planar-model-model-correction/simp-bike-input2.pdf}

    \caption{FML model  correction for simplified planar bicycle model~\eqref{eq:simp-bicyle} with slow ramp throttle with constant steering~\eqref{eq:control-slow-ramp} input signal; Top: true/prior/FML prediction trajectory and relative error for $x$ and $y$ over time; Bottom: true/prior/FML prediction and relative error for forward velocity and yaw angle  over time.}
    \label{Simp-Bicycle-correct2}
\end{figure}

\begin{figure}
  \includegraphics[width=\linewidth]{CSV/slip-free-planar-model-model-correction/simp-bike-input3.pdf}

    \caption{FML model  correction for simplified planar bicycle model~\eqref{eq:simp-bicyle} with smooth Gaussians~\eqref{eq:control-gauss} input signal; Top: true/prior/FML prediction trajectory and relative error for $x$ and $y$ over time; Bottom: true/prior/FML prediction and relative error for forward velocity and yaw angle  over time.}
    \label{Simp-Bicycle-correct}
\end{figure}
\end{comment}

\subsection{Full Planar Bicycle Model with Slip}
Now consider~\eqref{eq:full-bicyle} with initial conditions:$x(0)=1,y(0)=1,\psi(0)=1,v_x(0)=1,v_y(0)=0,\omega(0)=0$,  we test on the following new control input settings:
\begin{align}
    &\text{(a) High frequency steering}:
    \begin{cases}\label{eq:control-high-frq}
        u(t)=0.2+0.2\cos(0.3t), \\
        \delta(t)=0.5\sin(t),
    \end{cases} \\
    &\text{(b) Pulse acceleration and steering}:
    \begin{cases}\label{eq:control-pulse-full}
        u(t)=0.4\sin^2(0.5t), \\
        \delta(t)=0.02\cos^2(0.25t),
    \end{cases} \\
    &\text{(c) Piecewise trigonometric composition}:
    \begin{cases}\label{eq:control-pw-tri}
        u(t)=0.4\tanh(0.5t), \\
        \delta(t)=\frac{0.3\sin(0.5t)}{0.1t+1}.
    \end{cases}
\end{align}

\subsubsection{Learning Full Planar Bicycle Model with Slip via FML}\label{sec:full-bicyle-fml}
This subsection examines Case I, where the prior model~\eqref{eq:full-bicyle} is assumed to perfectly represent the true dynamics, with the true parameter values $\gamma_\text{true} = \gamma_\text{prior} = [b_u,b_\delta,L_f,L_r,m,I_z,C_f,C_r] = [5.0,0.4,0.082, 0.098, 2.5, 0.015, 2.0,2.0]$. Due to the system’s strong nonlinearity and complexity, it is challenging to construct an appropriate lifting map that enables DRIPS to achieve satisfactory performance. Consequently, we adopt FML as the learning framework, leveraging its higher expressivity at the cost of increased data requirements.

To construct the training dataset, we first fixed the initial condition as
$\mathbf{s}_0 = (1,1,1,1,0,0)$. We consider control trajectories 
as the following cosine functions:
\begin{equation}\label{eq:training_signals}
    u(t) = a_u + A_u \cos(\omega_u t + \phi_u), \qquad 
    \delta(t) = a_\delta + A_\delta \cos(\omega_\delta t + \phi_\delta),
\end{equation}
where the parameters are drawn uniformly so that 
$u(t) \in [0,\,0.8]$ and $\delta(t) \in [-0.6,\,0.6]$ for all $t \in [0,5]$, hence the coefficient space is taken to be $\Omega_\mathbf{p} =[0,0.8] \times [-0.6,0.6]$.
This ensures that the input signal training data remains within  
the control bounds of interest. The parameterization of the training test signals in Eq.~\eqref{eq:training_signals} are sampled uniformly from the following domains:
\begin{equation}
\begin{aligned}
\omega_u      &\sim \mathcal{U}[0.25,\,1.05],  &
\omega_\delta &\sim \mathcal{U}[0.45,\,1.05], \\[2pt]
a_u           &\sim \mathcal{U}[0.10,\,0.45],  &
a_\delta      &\sim \mathcal{U}[-0.01,\,0.05], \\[2pt]
A_u           &\sim \mathcal{U}[0.10,\,0.45],  &
A_\delta      &\sim \mathcal{U}[0.003,\,0.55], \\[2pt]
 \phi_u &\sim \mathcal{U}[0,\,2\pi], &  \phi_\delta &\sim \mathcal{U}[0,\,2\pi].
\end{aligned}
\end{equation}

The training data input–output pairs are constructed by first solving the ODE system up to time 
$T = 5$ under randomly generated control signals for $4000$ trajectories. From each trajectory, 
$20$ consecutive state variable pairs are randomly selected, resulting in a total training set 
of $J = 80{,}000$ data pairs sampled from $\Omega_{\mathbf{s}} \times \Omega_{\mathbf{p}}$. 
The state space domain is taken to be the following:
\[
\Omega_{\mathbf{s}} = [0,25] \times [0,28] \times [0.6,1.5] 
\times [0,15] \times [-2,1] \times [-0.6,0.6].
\]
We take the network architecture to be 3 hidden layers, 100 neurons per layers and train the network for 20,000 epochs with the ADAM optimizer and a triangular cyclic learning-rate schedule varying from $10^{-4}$ to $3 \times 10^{-3}$.

\begin{figure}
    \centering
    \includegraphics[width=0.9\linewidth]{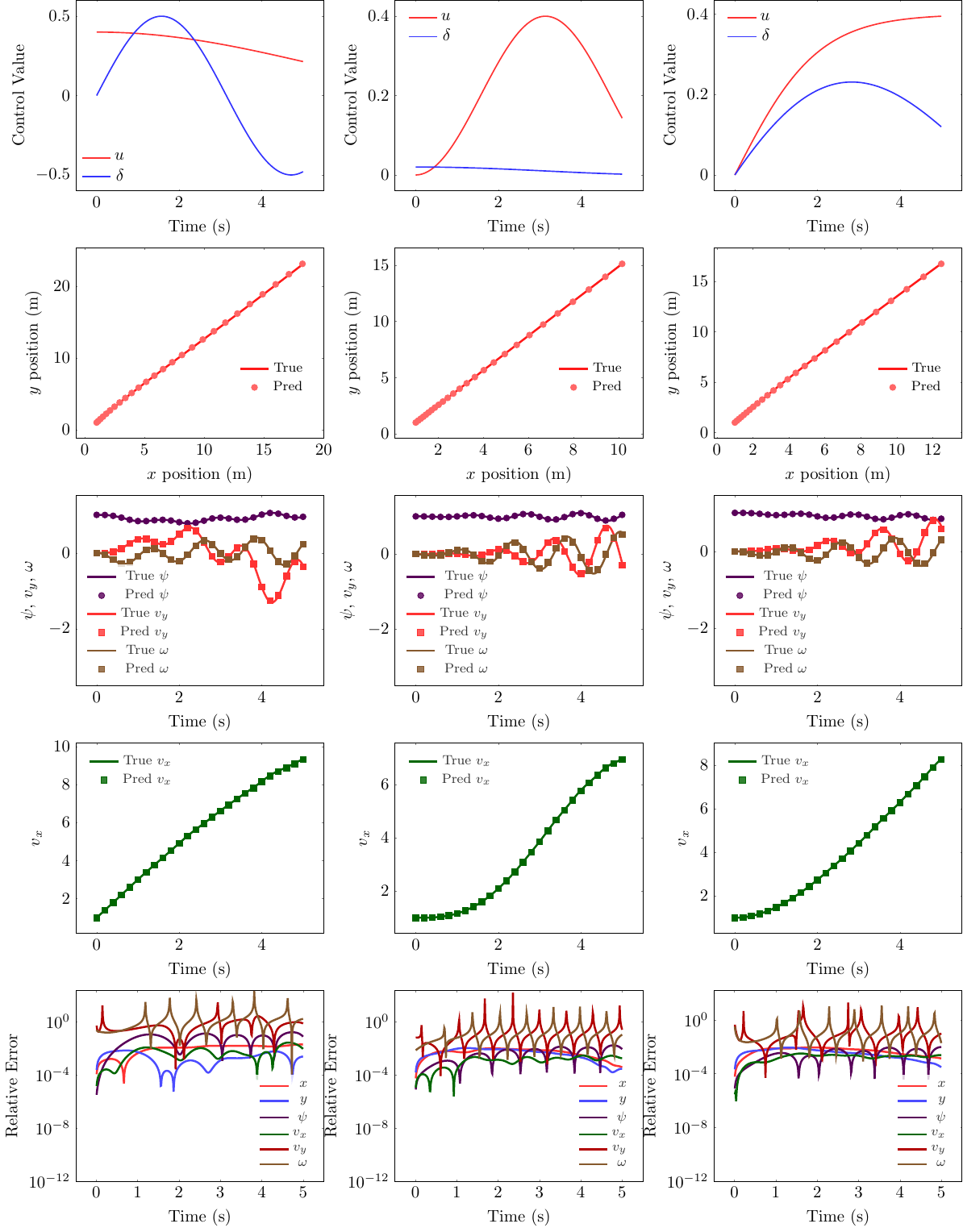}

    \caption{FML for the full planar bicycle model with slip ~\eqref{eq:full-bicyle} with testing control inputs: Left: (a) high frequency steering~\eqref{eq:control-high-frq}; Middle: (b) pulse acceleration and steering ~\eqref{eq:control-pulse}; Right: (c) piecewise trigonometric composition~\eqref{eq:control-pw-tri}. Row 1: control inputs over time; Row 2: true/FML prediction trajectory; Row 3: true/FML prediction forward velocity and yaw angle over time; Row 4: prediction error of the FML model, as defined in~\eqref{eq:modi-l2-err}.}
    \label{fig:slip1_t5}
\end{figure}

We present results for the three control input settings in Figure~\ref{fig:slip1_t5}; note the test control input settings are not included in the training dataset. We demonstrate good agreement between the trained flow-map network prediction and the true underlying trajectory for the three test signals. %In Figure \ref{fig:slip1_t10}, we further demonstrate the ability of the network to conduct accurate predictions up to time $T=10$ for the high frequency steering signal and thereby the ability of the flow map to extrapolate beyond the computational domain.  

% \begin{figure}
%     \centering
%     \includegraphics[width=\linewidth]{figures/final_bicycle_w_slip/slip_bicycle_combined1_T10.png}

%     \caption{Test case 1. T=10.}
%     \label{fig:slip1_t10}
% \end{figure}

\subsubsection{Model Correction}\label{sec:correction-full}
\begin{figure}
    \centering
    \includegraphics[width=0.9\linewidth]{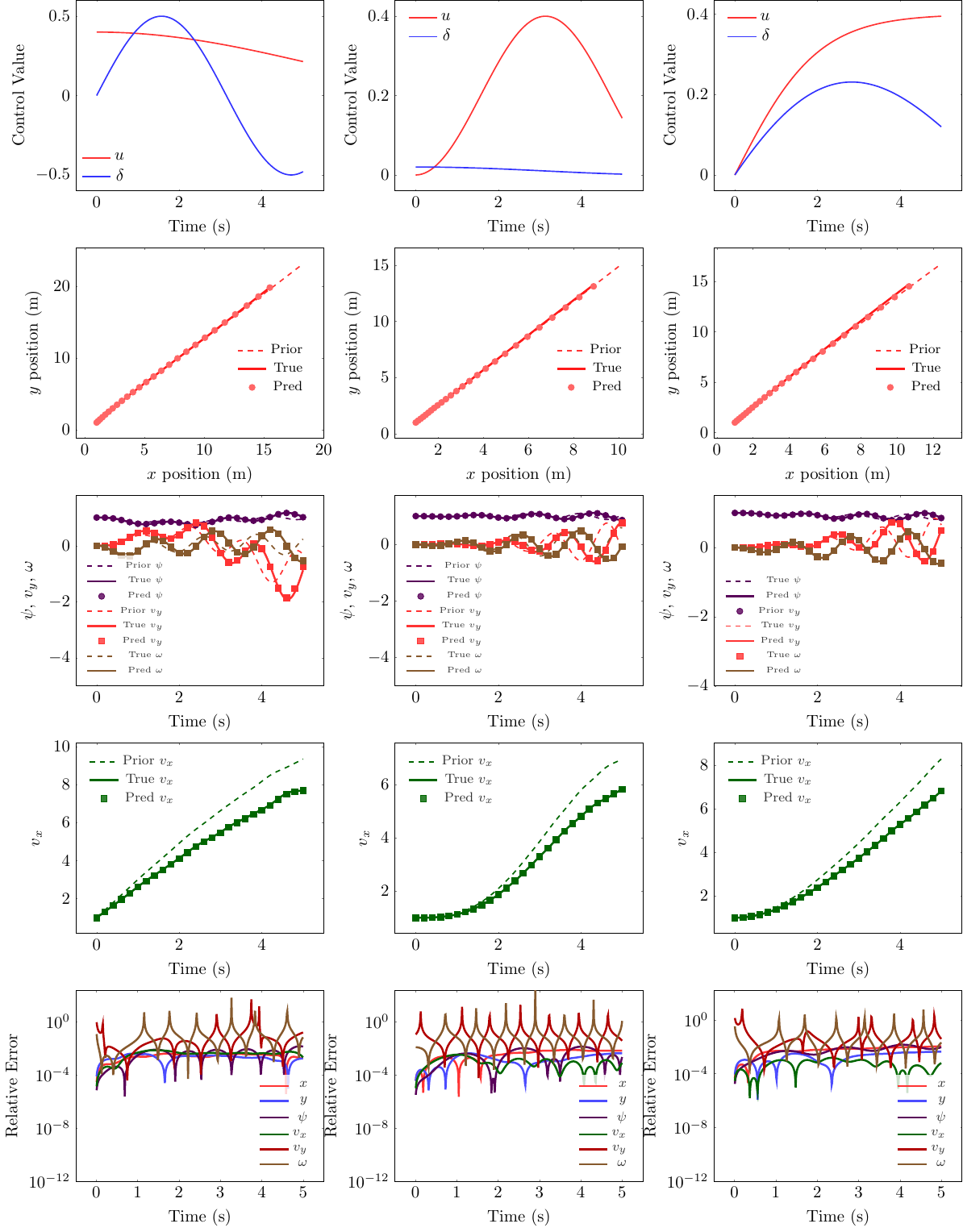}
    \caption{FML for model correction of the full planar bicycle model with slip ~\eqref{eq:full-bicyle} with testing control inputs: Left: (a). high frequency steering~\eqref{eq:control-high-frq}; Middle: (b). pulse acceleration and steering ~\eqref{eq:control-pulse}; Right: (c). piecewise trigonometric composition~\eqref{eq:control-pw-tri}. Row 1: control inputs over time; Row 2: true/prior/FML prediction trajectory; Row 3: true/prior/FML prediction forward velocity and yaw angle over time; Row 4: prediction error (as defined in~\eqref{eq:modi-l2-err}) of the FML model.}
    \label{fig:slip_corrected_signal1}
\end{figure}

We now perform a Case II example for model correction, where we consider the prior model with the parameterization described in the previous section. The true parameterization uses different values for $L_f$ and $L_r$, the distances from the center of mass to the front and rear wheels, respectively, as well as a different scaling coefficient $b_u$. Whereas the prior model considers $[L^{\rm prior}_f, L^{\rm prior}_r] = [0.082, 0.098]$ m and $b_u^{\rm prior} = 5$, the true parameter values are taken to be $[L^{\rm true}_f, L^{\rm true}_r] = [0.10, 0.05]$ and $b_u^{\rm true} = 5$. We freeze the first two layers of the network and conduct transfer learning on the final layer by minimizing Eq.~\eqref{eq:NN_par_HF}; the last two layers are trained for up to 5000 epochs using ADAM with a learning rate of $10^{-3}$ and early stopping (patience of 100). The scarce high-fidelity data set consists of 10 randomly selected input-output pair samples from 100 trajectories, each of length $T=5$, giving a total of $J_{HF} = 1000$ data pairs. The input signal parameters for training are sampled as described in the previous section. Results presented in Figure~\ref{fig:slip_corrected_signal1} show good agreement between the corrected flow map model predictions and the true solutions for all control inputs. 

\subsection{Model Correction with Experimental Data}\label{sec:exp-data}
Finally, we examine Case III, applying our model-correction methods in a real-world setting using a small-scale autonomous vehicle—the NVIDIA JetRacer Tamiya TT-02. As our initial prior model $\textbf{f}_{\rm prior}$ of the vehicle dynamics, we consider the following modified system of the simplified planar bicycle dynamics in Eq.~\eqref{eq:simp-bicyle}:

\begin{equation}\label{eq:prior-experimental}
    \begin{aligned}
        & \frac{d}{dt} x(t) =v_x(t)\cos(\psi(t)) \\
        &\frac{d}{dt}y(t)=v_x(t)\sin(\psi(t)) \\
        &\frac{d}{dt}v_x(t)=
        \begin{cases}
            b_u u(t), \ &u(t)\geq0.13 \\
            0,&u(t) < 0.13
        \end{cases} \\
        &\frac{d}{dt}{\psi(t)}=\frac{v_x(t)}{L}\tan(b_\delta\delta(t))
    \end{aligned}
\end{equation}
where we fix parameters $b_u=4.6$, $L = 0.255$ and $b_\delta=1.35$.

During the experiments, the control inputs were defined as
\begin{subequations}\label{eq:control-input-experimental}
\begin{align}
\begin{cases}
    u(t)= 0.05t\exp(0.05t)\\  
    \delta(t)=0.1. 
\end{cases}
\end{align}
\end{subequations}
and were applied to the vehicle at 100 Hz. The states $[x(t), y(t), v_X(t), v_Y(t), \yaw(t)]^T$ were initialized using the test initial condition 
$$[x(0), y(0), v_X(0), v_Y(0), \yaw(0)]^T = [2.818, 2.887,0.027,0.017,-3.133]^T.$$ 
Trajectory data were measured by a motion-capture system operating at the same sampling rate in an inertial frame. The values $(v_X(t), v_Y(t))$ denote the inertial world frame velocities, which are post-processed by applying a yaw-axis rotation transformation for determining the body-frame forward velocity $v_x$:
\begin{equation}\label{eq:rotation}
    v_x(t)=v_X(t)\cos(\psi(t))+v_Y(t)\sin(\psi(t)).
\end{equation}
Our state values of interest are then taken to be $\mathbf{s}(t)= [x(t), y(t), v_x(t), \yaw(t)]^T$. 

Data were collected at a constant timestep of $\Delta t = 0.01$ seconds over a time horizon of $[0,T]$ for $T=5$ seconds and post-processing smoothing is applied using a moving-average filter. Hence a total of $J_{HF} = 500$ high-fidelity input-output data pairs are collected to form $\mathcal{S}_{\rm train}$. 

To train the prior flow map, the computational domain is taken to be \(\Omega_\mathbf{\mathbf{s}} = [0,4] \times [-1,4] \times [-1,3] \times [-4,-1]\). To approximate the one-step evolution operator, we use quadratic polynomials for local parameterization of signal $u(t)$ where the coefficient domain is taken to be \(\Omega_{\mathbf p}= [0,0.4]^3\). We consider a fixed value for our second input control, setting  $\delta(t) = 0.1$ for all $t$. To form $\mathcal S_{\text{prior}}^{\text{train}}$, we uniformly sample from $\Omega_{\mathbf{s}} \times \Omega_{\mathbf{p}}$ and generate $J_{LF} = 50{,}000$ input–output pairs by solving the prior system in Eq.~\eqref{eq:prior-experimental}.

\begin{figure}
    \centering
    \includegraphics[width=\linewidth]{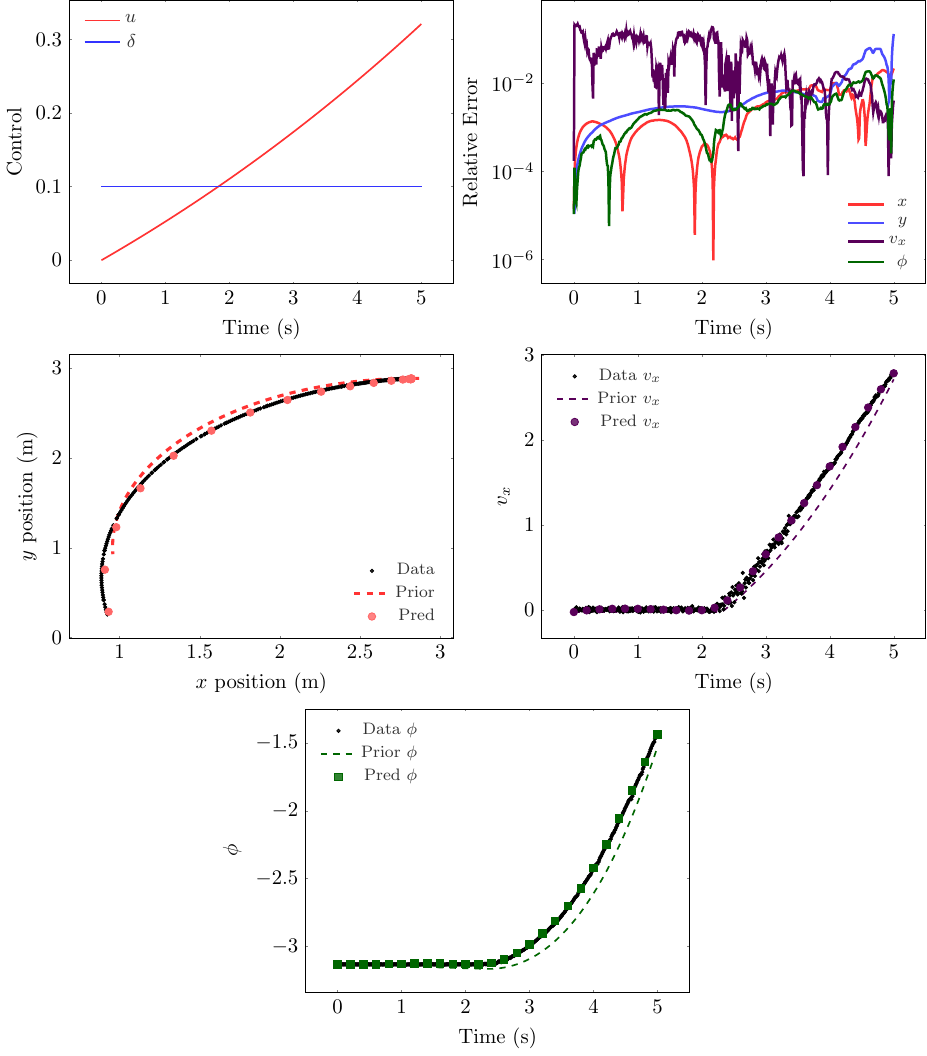}
    \caption{FML for model correction of the simplified planar bicycle model ~\eqref{eq:prior-experimental} using experimental data. We test using the control input defined in ~\eqref{eq:control-input-experimental}. Row 1: control inputs over time and relative error of the corrected FML model against the ground truth smoothed observed states; Row 2 and 3: The true/prior/FML prediction trajectory of states over $[0,5]$.}
    \label{fig:Experimental-Simp-Bicycle-fml}
\end{figure}

The DNN is taken to be 3 hidden layers, each containing 100 nodes. The network is trained using  $\mathcal{S}_\text{prior}^\text{train}$, training for $10^3$ epochs using the ADAM optimizer and a learning rate of $10^{-3}$. For model correction, we utilize the experimental data set $\mathcal{S}_{\rm train}$ of $J_{HF} = 500$ data pairs. Here conduct transfer learning on all layers, using the loss function defined in~\eqref{eq:NN_par_HF}; we again train for at most 5000 epochs with early stopping (patience of 100) to obtain the corrected flow map $\widetilde{\mathbf{N}}_{\Theta^*}$. We demonstrate results for the test initial condition in Figure~\ref{fig:Experimental-Simp-Bicycle-fml}; we observe that after applying our correction method, we are now able to approximate the experimentally observed dynamics for all state variables with good accuracy. We note that our model correction method does not assume a specific form of the model correction terms, and by leveraging the flow-map of the prior simplified planar bicycle model, we are able to correct the complex and unknown model-form error of the experimentally observed vehicle dynamics.

\section{Conclusion}\label{sec:conclusion}

In this work, we developed a data-driven framework for learning and correcting non-autonomous vehicle dynamics, enabling accurate predictions even when only limited high-fidelity or experimental data are available. By locally parameterizing the time-dependent control inputs, we transformed the original non-autonomous dynamics into a sequence of locally parametric autonomous systems, on which DRIPS and FML can be applied effectively. Our results demonstrate that DRIPS yields highly data-efficient surrogate models, while FML, augmented with a transfer-learning-based correction mechanism, can refine imperfect physics-based models without assuming an explicit additive or multiplicative correction form and without requiring large quantities of high-fidelity data. This framework, therefore, bridges idealized physics-based models with real-world measurement data, improving predictive fidelity while retaining interpretability. Numerical experiments across multiple vehicle models verify that the corrected surrogates achieve high prediction accuracy even under strong nonlinearity and model-form error. Looking ahead, this approach offers a promising foundation for compositional surrogate modeling of multi-vehicle systems, where locally learned or corrected data-driven models can be fused to capture coupled interactions, enabling scalable prediction and control in complex, multi-agent environments.

\bibliographystyle{plain}
\bibliography{main}
\newpage
\section*{Appendix}
\label{app:lifted_dynamics}

We define the lifted system equation 
appearing in Section~\ref{sec:sim-bicyle-drips} as
\begin{equation}
    \frac{d\mathbf{g}(t)}{dt}
    = A\bigl(u(t),\delta(t);L,b_\delta,b_u\bigr)\mathbf{g}(t)
      + b\bigl(u(t),\delta(t);L,b_\delta,b_u\bigr).
\end{equation}
where

\footnotesize  % or \footnotesize or \scriptsize if you need more shrink

\begingroup
\setlength{\arraycolsep}{3pt} % tighten column spacing
\footnotesize
\begin{equation}
A(u,\delta) =
{\renewcommand{\arraystretch}{2.0}
\begin{bmatrix}
  0 & 0 & 0 & 0 & 0 & 0 & 1 & 0 & 0 & 0 & 0 & 0 \\
  0 & 0 & 0 & 0 & 0 & 0 & 0 & 1 & 0 & 0 & 0 & 0 \\
  0 & 0 & 0 & 0 & 0 & 0 & 0 & 0 & 0 & 0 & 0 & 0 \\
  0 & 0 & 0 & 0 & 0 & 0 & 0 & 0 & 0 & 0 & 0 & 0 \\
  0 & 0 & 0 & 0 & 0 & 1 & 0 & 0 & 0 & 0 & 0 & 0 \\
  0 & 0 & 0 & 0 & 1 & 0 & 0 & 0 & 0 & 0 & 0 & 0 \\
  0 & 0 & 0 & 0 & b_u u & 0 & 0 & -\dfrac{\tan(b_\delta\delta)\!\int b_u u}{L} & 0 & 0 & 0 & 0 \\
  0 & 0 & 0 & 0 & 0 & b_u u & \dfrac{\tan(b_\delta\delta)\!\int b_u u}{L} & 0 & 0 & 0 & 0 & 0 \\
  0 & 0 & 0 & 0 & 0 & 0 & 0 & 0 & 0 & 0 & 0 & 0 \\
  0 & 0 & 0 & 0 & 0 & 0 & 0 & 0 & 0 & 0 & 0 & -\dfrac{2\tan(b_\delta\delta)\!\int b_u u}{L} \\
  0 & 0 & 0 & 0 & 0 & 0 & 0 & 0 & 0 & 0 & 0 & \dfrac{2\tan(b_\delta\delta)\!\int b_u u}{L} \\
  0 & 0 & 0 & 0 & 0 & 0 & 0 & 0 & 0 & \dfrac{\tan(b_\delta\delta)\!\int b_u u}{L} & -\dfrac{\tan(b_\delta\delta)\!\int b_u u}{L} & 0
\end{bmatrix}}
\end{equation}
\endgroup

\begin{equation}
b(u,\delta) =
\begin{bmatrix}
    0 \\ 0 \\ b_u u \\ \dfrac{\tan(b_\delta\delta)\!\int b_u u}{L} \\
    0 \\ 0 \\ 0 \\ 0 \\ 2 b_u u \int b_u u \\ 0 \\ 0 \\ 0
\end{bmatrix}.
\end{equation}

\end{document}